\documentclass{article}
\usepackage{graphicx} 
\usepackage{multirow}
\usepackage{graphicx} 
\usepackage{amsmath}
\usepackage[table]{xcolor} 
\usepackage{authblk}  
\usepackage{xcolor}
\usepackage{soul}
\usepackage{url}
\usepackage{colortbl} 
\usepackage{amssymb} 
\usepackage{pifont}  
\usepackage{geometry} 
\usepackage{pdflscape} 
\usepackage{algorithm}
\usepackage{threeparttable} 
\usepackage[font=small]{caption}
\usepackage{algpseudocode}
\usepackage{notoccite} 
\usepackage{graphicx}
\usepackage{subcaption}
\usepackage{float}
\usepackage{caption} 
\sethlcolor{yellow} 

\title{ALCM: Autonomous LLM-Augmented Causal Discovery Framework}

\date{} 

\author[1]{Elahe Khatibi}
\author[1]{Mahyar Abbasian}
\author[1]{Zhongqi Yang}
\author[1]{Iman Azimi}
\author[1,2]{Amir M. Rahmani}

\affil[1]{\small Department of Computer Science, University of California, Irvine, USA}
\affil[2]{\small School of Nursing, University of California, Irvine, USA}

\begin{document}

\maketitle

\section*{Abstract}

To perform effective causal inference in high-dimensional datasets, initiating the process with causal discovery is imperative, wherein a causal graph is generated based on observational data. However, obtaining a complete and accurate causal graph poses a formidable challenge, recognized as an NP-hard problem. Recently, the advent of Large Language Models (LLMs) has ushered in a new era, indicating their emergent capabilities and widespread applicability in facilitating causal reasoning across diverse domains, such as medicine, finance, and science. The expansive knowledge base of LLMs holds the potential to elevate the field of causal reasoning by offering interpretability, making inferences, generalizability, and uncovering novel causal structures. In this paper, we introduce a new framework, named Autonomous LLM-Augmented Causal Discovery Framework (ALCM), to synergize data-driven causal discovery algorithms and LLMs, automating the generation of a more resilient, accurate, and explicable causal graph. The ALCM consists of three integral components: causal structure learning, causal wrapper, and LLM-driven causal refiner. These components autonomously collaborate within a dynamic environment to address causal discovery questions and deliver plausible causal graphs. We evaluate the ALCM framework by implementing two demonstrations on seven well-known datasets. Experimental results demonstrate that ALCM outperforms existing LLM methods and conventional data-driven causal reasoning mechanisms. This study not only shows the effectiveness of the ALCM but also underscores new research directions in leveraging the causal reasoning capabilities of LLMs.

\textbf{Keywords}: Large Language Models; Causal Reasoning; Causal Graph; Causal Discovery;

\section{Introduction}


 The process of causal discovery, essential in various domains and scientific discoveries, seeks to reveal complex causal relationships in observational data \cite{pearl2009causal, pearl2012causal, glymour2019review}. For instance, in healthcare, this process is crucial and instrumental for pinpointing disease etiologies, devising effective interventions, and prevention strategies \cite{zanga2022survey}. Subsequently, causal inference allows for the quantification of the influence exerted by different variables on one another, once a causal structure is identified. This phase, often referred to as causal estimation, relies on the construction of a preliminary causal graph, which, despite its theoretical significance, poses considerable practical challenges, demanding substantial domain-specific expertise. In fact, studies using real-world datasets demonstrate that inferring causal graphs--which is the focus of this paper--from data is still a complex challenge in practical applications \cite{reisach2021beware, tu2019neuropathic, ban2023causal}. Causal discovery and causal inference, as highlighted in seminal works by Pearl and others \cite{pearl2009causal, pearl2012causal, liu2024discovery, glymour2019review}, are two key components of causal reasoning to address causal questions in diverse fields.


Within the literature, numerous studies have contributed significantly to the development of a variety of efficient causal discovery algorithms aimed at uncovering the underlying causal structure from observational data. This body of research can be broadly categorized into two main groups: conventional data-driven causal discovery algorithms and those based on LLMs \cite{wan2024bridging}.
Conventional causal discovery algorithms focus on learning the causal graph from samples of the joint probability distribution of observational data. They utilize various statistical techniques, including conditional independence tests, machine learning approaches, generative models, deep learning methodologies, and reinforcement learning strategies \cite{sauter2023meta} to understand the joint distribution of observed variables and extract the causal connections among them. Subsequently, these algorithms assess how well the candidate causal graph aligns with the data \cite{zanga2022survey, hasan2023survey, glymour2019review}.



Conventional causal discovery algorithms, despite being designed to be powerful and scalable, face several challenges. These include a heavy dependence on domain experts \cite{constantinou2023impact}, who are often limited and inconsistent, and the issues of data bias, imbalance, and inadequacy which affect the accuracy of capturing true probability distributions \cite{ban2023causal}. Additionally, the use of static data can compromise model accuracy in dynamic environments, and the task of fully determining edge orientations is hindered by the presence of multiple equivalent Directed Acyclic Graphs (DAGs) \cite{ban2023causal, sauter2023meta}, which exponentially increase with the number of nodes \cite{zheng2018dags}, leading to inaccuracies and unreliability in the estimated causal graphs.

Recent advancements in Large Language Models (LLMs) have significantly impacted artificial intelligence, exhibiting notable reasoning capabilities \cite{kiciman2023causal, wang2024exploring, cai2023knowledge, lampinen2023passive, antonucci2023zero}. These achievements stem from the extensive data used for training LLMs, essential for effective causal reasoning \cite{kiciman2023causal, cai2023knowledge}. However, current LLM-based causal reasoning research, mainly focusing on pairwise analysis, faces scalability issues as it struggles with the complexity of full causal graph construction and handling large datasets \cite{vashishtha2023causal, kiciman2023causal, ban2023causal, ban2023query, pawlowski2023answering}. These models often fall short in accurately and efficiently inferring comprehensive causal relationships, especially in large-scale settings \cite{ban2023query, ban2023causal, liu2024discovery, jiralerspong2024efficient}. Despite some efforts to integrate LLMs with causal discovery processes \cite{vashishtha2023causal, ban2023query, takayama2024integrating}, challenges remain due to inherent limitations and the complexity of causal inference. A synergistic approach combining LLMs with other methods may provide a more nuanced and complete understanding of causal mechanisms and address these challenges effectively.

In this paper, we present an LLM-powered causal discovery framework--ALCM: a multi-component Autonomous LLM-Augmented Causal Discovery Framework. ALCM proposes a synergized reasoning method and entails three components: causal structure learning, causal wrapper, and LLM-driven refiner components to generate more accurate and robust causal graphs. ALCM is engineered to autonomously untangle causal structures by deciphering those causal relations embedded in observational data. ALCM capitalizes on observed data, data-driven causal reasoning algorithms, and the implicit knowledge embedded in LLMs to optimize and streamline the entire causal reasoning process. This approach aims to establish a more robust, applicable, and reliable foundation for causal reasoning and estimation as well. We conduct a comprehensive performance evaluation of ALCM, employing LLMs and assessing their capabilities on widely recognized benchmarks \cite{bnlearnBnlearnBayesian, tu2019neuropathic}. We compare our framework with conventional causal discovery algorithms and LLMs prompting. Furthermore, we implement an automatic pipeline for making the causal discovery an automatic task. Our contributions are as follows:

Our contributions in this work are as follows:

\begin{itemize}
    \item \textbf{Unified Framework for Enhanced Causal Discovery:} We introduce the ALCM framework that synergistically integrates the strengths of conventional data-driven causal discovery (CCD) methods and Large Language Models (LLMs) to overcome the limitations of individual approaches by generating accurate, interpretable, and comprehensive causal graphs.
    
    \item \textbf{Dynamic, Scalable, and Autonomous Operations:} ALCM demonstrates adaptability to dynamic data environments, autonomous operation without domain expertise dependency, and scalability across unseen datasets. 
    
    \item \textbf{Improved Predictive Precision and Reliability:} Leveraging the contextual reasoning capabilities of LLMs, the framework refines causal relationships with algorithmic rigor. 
    
    \item \textbf{Comprehensive Graph Representation and Explainability:} ALCM provides a fully automated pipeline for constructing and refining causal graphs. It ensures interpretability and explainability of results.
    
    \item \textbf{Benchmarking and Performance Validation:} We extensively evaluate ALCM across multiple benchmark datasets, demonstrating its superior performance compared to the related works. 
\end{itemize}


This work advances the field of causal discovery by demonstrating the transformative potential of a unified framework that synthesizes the complementary capabilities of CCD and LLM-based approaches.

\section{Background and Related Work}

In this section, we outline the existing research on causal structure learning within the literature, delineating it into two primary groups: 1) Conventional data-driven causal discovery algorithms; and 
2) Using LLMs for causal discovery. 

 \textbf{1) Conventional data-driven causal discovery algorithms}: conventional data-driven causal discovery algorithms are broadly classified into five categories as follows:

\begin{itemize}
    \item \textbf{Score-Based Algorithms}: They operate on scores and engage in a comprehensive exploration of the entire space of potential Directed Acyclic Graphs (DAGs) to identify the most suitable graph for explaining the underlying data. Typically, such score-based approaches consist of two integral components: (i) a systematic search strategy tasked with navigating through the potential search states or the space of candidate graphs, denoted as G', and (ii) a score function designed to evaluate the viability of these candidate causal graphs. The synergy between the search strategy and the score function is instrumental in optimizing the exploration of all conceivable DAGs. A widely employed score function in the selection of causal models is the Bayesian Information Criterion (BIC) \cite{hasan2023survey}. Some examples of score-based algorithms are Greedy Equivalence Search (GES) \cite{chickering2015selective}, Fast Greedy Search (FGS) \cite{ramsey2015scaling}, and A* Search \cite{xiang2013lasso}.

\item \textbf{Constraint-Based Algorithms}: This category, exemplified by Peter-Clark (PC) \cite{spirtes2000causation} algorithm, employs conditional
independence (CI) tests to reveal the graph’s skeleton and v-structures, ultimately returning the Directed Acyclic
Graph (DAG) of the functional causal model while considering v-structures and doing edge-orientations \cite{hasan2023survey}. Other constraint-bsaed algorithms are like Fast Causal Inference (FCI), Anytime FCI, RFCI, PC-stable, and so forth. 

\item \textbf{Hybrid Algorithms}: Hybrid approaches are founded on the integration of various causal discovery methods, combining constraint-based, score-based, Functional Causal Model (FCM)-based, gradient-based, and other techniques. This amalgamation reflects a comprehensive strategy that leverages the strengths of different methodologies to enhance the robustness and effectiveness of causal discovery in complex systems. Max-Min Hill Climbing (MMHC) \cite{tsamardinos2006max}--belonging to this category--stands out as a hybrid causal discovery technique that seamlessly integrates principles from both score-based and constraint-based algorithms. This hybrid approach combines the advantages of scoring methods and constraint-based strategies, offering a comprehensive and effective framework for uncovering causal relationships in complex systems.

\item \textbf{Function-Based Algorithms}: 
Approaches grounded in Functional Causal Models (FCM) delineate the causal connections between variables within a defined functional structure. In FCMs, variables are expressed as functions of their direct causes (parents), augmented by an independent noise term denoted as E. The distinguishing feature of FCM-based methodologies lies in their capacity to differentiate between various Directed Acyclic Graphs (DAGs) within the same equivalence class. This discrimination is achieved by introducing supplementary assumptions concerning data distributions and/or function classes. Several notable FCM-based causal discovery methodologies are introduced, including Linear Non-Gaussian Acyclic Model (LiNGAM) \cite{shimizu2006linear} and  Structural Agnostic Modeling (SAM) \cite{kalainathan2022structural}. SAM employs an adversarial learning methodology for causal graph identification. Specifically, SAM utilizes Generative Adversarial Neural Networks (GANs) to seek a Functional Causal Model (FCM) while ensuring the detection of sparse causal graphs through the incorporation of appropriate regularization terms. The optimization process involves a learning criterion that integrates distribution estimation, sparsity considerations, and acyclicity constraints. This holistic criterion facilitates end-to-end optimization of both the graph structure and associated parameters, accomplished through stochastic gradient descent.

The previous three-mentioned categories may be limited to the Markov equivalence class, posing
constraints. Function-based algorithms like LiNGAM [44] aim to uniquely identify causal DAGs by exploiting data
generative process asymmetries or causal footprints.

\item \textbf{Optimization-Based Algorithms}: Recent investigations in causal discovery have approached the structure learning problem by casting it as a continuous optimization task, employing the least squares objective and an algebraic representation of Directed Acyclic Graphs (DAGs). Notably, this transformation converts the combinatorial nature of the structure learning problem into a continuous framework, and solutions are obtained through the application of gradient-based optimization techniques. These methods exploit the gradients of an objective function concerning the parameterization of a DAG matrix to achieve effective structure learning. NOTEARS \cite{zheng2018dags} is among the causal discovery algorithms that formulate the structure learning problem as a purely continuous constrained optimization task.

\end{itemize}

\textbf{2) Using LLM for causal discovery task}: Leveraging recent advancements in LLMs and Natural Language Processing (NLP) presents an opportunity to offer enhanced capabilities in capturing causal concepts and relations while handling large-scale datasets more effectively \cite{long2023can, chen2023mitigating, nori2023capabilities}. This proficiency is rooted in the extensive training
LLMs undergo on vast, high-quality datasets [18]. LLMs possess the ability to establish a comprehensive knowledge base across diverse domains, facilitating language understanding, ensuring
generalizability, automating the causal reasoning pipeline, and enabling plausible reasoning. In this regard, the second group, namely using LLMs for causal discovery, is introduced. This group is classified into three major groups as follows:
\begin{itemize}
    \item \textbf{Fine-tuning}: This category mainly focuses on fine-tuning LLMs to empower LLMs with causal-and-effect knowledge and address the causal reasoning challenges \cite{jin2023can, abdulaal2023causal, jiang2023large}. For instance, Jin et al. \cite{jin2023can} introduce the CORR2CAUSE benchmark dataset on which they fine-tune their model. This is done to both asses and empower LLMs with causal reasoning ability. In fact, CORR2CAUSE dataset serves as a tool to evaluate the proficiency of LLMs in discerning causal relationships, particularly when the LLMs are fine-tuned to distinguish causation from correlational statements in the context of NLP.

\item \textbf{Performance Evaluation}: The second category focuses on using LLM for causal discovery and delves into emerging research that explores the causal analysis capabilities of Large Language Models. In contrast to causal discovery algorithms relying on statistical patterns in the data, this group utilizes LLMs to discover causal structures from variables. A majority of these methods solely utilize LLMs to predict pairwise causal relationships among a given set of variables \cite{willig2022can, liu2023evaluating, kiciman2023causal, tucausal, pawlowski2023answering, ban2023causal, zhang2024causal}.  


\item \textbf{Prior or Posterior Knowledge}: In the third category, focused on employing LLMs, the objective is either to assign direction to undirected edges generated by causal discovery algorithms or to impose constraints on the edge orientation and functionality of these algorithms.

 \cite{ban2023query, ban2023causal, vashishtha2023causal}.
\end{itemize}

Despite these efforts from conventional data-driven causal discovery algorithms to propose robust, precise, adaptable, efficient, and scalable causal discovery algorithms, encountered limitations and inefficiencies persist. These challenges are as follows. 1) Real-world data, often sparse and insufficient for accurately capture authentic probability distributions \cite{ban2023causal}. 2) Sole reliance on pre-collected static data introduces accuracy risks, particularly when models must adapt to dynamic real-world data and unforeseen factors. 3) Inferring complete edge orientations from observed data is hindered by the existence of equivalent Directed Acyclic Graphs (DAGs) \cite{ban2023causal, sauter2023meta}. 4) Algorithm dependence on domain knowledge experts, who may be scarce, time/resource-intensive, or exhibit variable quality across domains \cite{constantinou2023impact}. 5) Traditional causal discovery algorithms fall short in answering user-submitted causal questions due to a lack of proficiency in language understanding and processing. These challenges collectively contribute to diminished accuracy, incompleteness, and unreliability in the estimated causal graph. 

On the other hand, significant advances have been made in utilizing LLMs for causal tasks. However, their inherent limitations in precision and complexity handling remain evident. These challenges are highlighted as follows. 1) LLMs inherently lack the precision necessary for accurately responding to complex, user-generated causal queries \cite{tucausal}. 2) LLMs are limited in their ability to dissect and comprehend nuanced causal concepts without additional data-driven causal reasoning algorithms. 3) There is a challenge in constructing complete causal graphs and unraveling intricate causal relations due to the oversimplified understanding of LLMs. 4) LLMs struggle with handling extensive datasets, often failing to capture the depth and variability within them. These issues collectively hinder the effectiveness of LLMs in accurately and reliably determining causal relationships. Consequently, data-driven causal reasoning algorithms assume a critical role in mitigating the limitations of LLMs in causal tasks, offering nuanced comprehension of causal concepts, unraveling intricate causal relations, constructing complete causal graphs, and handling extensive datasets.

In light of these considerations, a unified, comprehensive causal framework that integrates LLMs with data-driven conventional causal discovery algorithms is required. To address this need, we propose the development of ALCM. ALCM aims to enhance the robustness and accuracy of causal discoveries by leveraging the conventional causal discovery algorithms and LLMs. 

Table \ref{comparison among three approaches} indicates the capabilities of two distinct causal discovery methods—Conventional data-driven Causal Discovery (CCD), LLMs-based approaches, and ALCM framework—across essential functional attributes. Dynamic Data Adaptability {\cite{liu2024discovery, ashwani2024cause, zhang2024causal}} is the capability of a method to adjust to changing data, while Detection of Hidden Variables \cite{liu2024discovery, zhang2024causal} refers to identifying unobserved influencers within the dataset. Comprehensive Graph Model Representation \cite{ban2023causal} assesses the completeness of the depicted causal structure, and Predictive Accuracy \cite{kiciman2023causal, takayama2024integrating, liu2024discovery, vashishtha2023causal, tucausal, pawlowski2023answering, zhang2024causal} measures the success in forecasting the correct causal relations.
CCD methods are limited by their reliance on pre-defined statistical models as well as domain knowledge expert validation, lacking adaptability to dynamic data, generalizability \cite{kiciman2023causal, jang2023can} to unseen data, autonomy, and lack of accuracy. Similarly, while LLMs are adept at dynamicity of data, generalizability, and detecting hidden variables, they fall short in providing comprehensive graph model representations, interpretability, explainability, autonomy, and precision for causal discovery task. ALCM combining these strengths while enhancing user independence from expert validation \cite{kiciman2023causal} and interpretability \cite{bhattacharjee2024towards} in causal discovery.



\begin{table}[H]
\centering
\caption{Comparative Analysis of CCD, LLMs, and ALCM across Key Functional Attributes}
\label{comparison among three approaches}
\begin{threeparttable}
\begin{tabular}{|c|c|c|c|}
\hline
\textbf{Descriptive Attribute} & \textbf{CCD} \tnote{1} & \textbf{LLMs} & \textbf{ALCM} \\ \hline
Dynamic Data Adaptability & $\times$ & $\checkmark$ & $\checkmark$ \\ \hline
Detection of Hidden Variables & $\times$ & $\checkmark$ & $\checkmark$ \\ \hline
Comprehensive Graph Model Representation & $\checkmark$ & $\times$ & $\checkmark$ \\ \hline
Predictive Accuracy & $\times$ & $\times$ & $\checkmark$ \\ \hline
Autonomous Operation & $\times$ & $\times$ & $\checkmark$ \\ \hline
Generalizability to Unseen Data & $\times$ & $\checkmark$ & $\checkmark$ \\ \hline
Autonomous Expert Validation & $\times$ & $\checkmark$ & $\checkmark$ \\ \hline
Interpretability and Explainability & $\checkmark$ & $\times$ & $\checkmark$ \\ \hline
\end{tabular}
\begin{tablenotes}
\item[1] CCD methods often rely on pre-defined statistical models and assumptions about the data generation process.
\item[2] LLMs-based methods may utilize vast amounts of data and natural language processing to infer causal relationships, potentially incorporating domain expertise.
\item[3] ALCM synthesizes the strengths of both CCDs and LLMs to uncover causal connections.

\end{tablenotes}
\end{threeparttable}
\end{table}


\section{Proposed Framework}

\begin{figure}[H]
    \centering
    \includegraphics[width=0.95\textwidth]{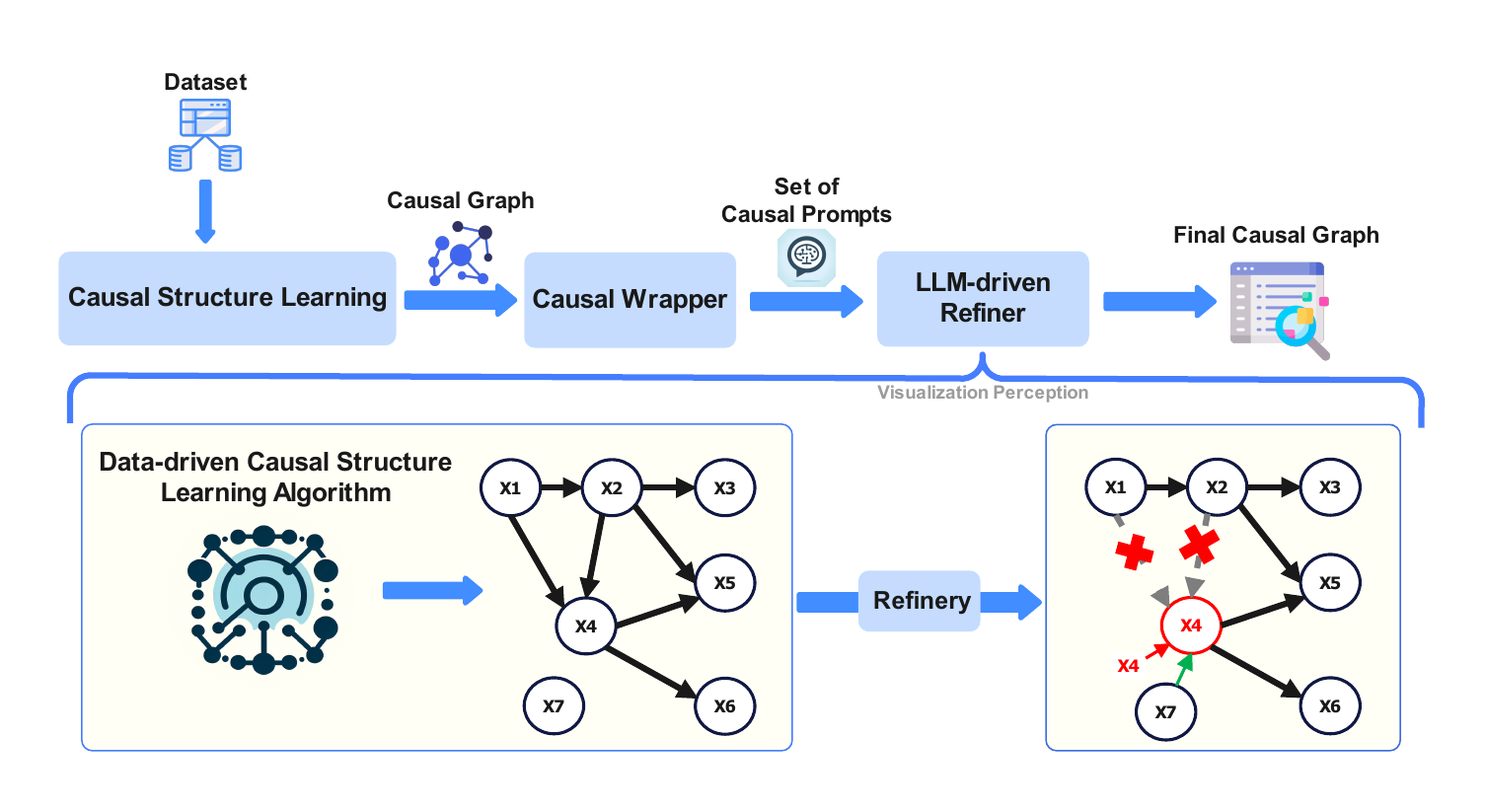}
    \caption{ALCM Architecture}
    \label{fig:ALCM Architecture}
\end{figure}

In this section, we present ALCM, an advanced causal discovery framework aimed to leverage the combined strengths of traditional causal discovery algorithms and LLMs. ALCM provide an automated pipeline constructing a comprehensive causal graph, refining it, and incorporating previously overlooked insights to enrich the resulting causal model. This integration aims to utilize the precision of conventional causal discovery algorithms in identifying data relationships, while also enhancing and validating these findings with insights from LLMs. Fig. \ref{fig:ALCM Architecture} indicates an overview of the ALCM framework. The algorithmic perspective of the ALCM framework is detailed in Algorithm \ref{alg:ALCM}. The ALCM framework includes three principal components: Causal Structure Learning, Causal Wrapper, and LLM-driven Refiner. These components interact iteratively to enhance the causal discovery process, to clarify the functionality and definitions of the framework, we present and exemplify these components and their interconnections in the following.  



\begin{algorithm}
\caption{ALCM}
\label{alg:ALCM}
\textbf{Require:} Observed dataset, $O$; Contextual Causal Information, $C$; Metadata, $M$ \\
\textbf{Ensure:} Causal DAG, $DAG$
\begin{algorithmic}[1]
\State \textbf{Initialize and run} the selected data-driven causal discovery algorithm, $CD$, to generate an initial causal graph $G_i \leftarrow CD(O)$ 
\Comment{Step 1: Generate the initial causal graph from observational data.}
\State \textbf{Generate} the causal prompt by injecting $C$ and $M$ 
\Comment{Step 2: Prepare contextual information for refinement by LLM.}
\For{each edge $z = (e_i, e_j)$ in $G_i$}
    \If{$z$ is validated by LLM-Driven Refiner} 
    \Comment{Step 3: Validate the existence of the causal edge using the LLM.}
        \State $G_i \gets G_i \cup \emptyset$ 
        \Comment{If valid, retain the edge.}
    \EndIf
    \If{$z$ orientation is revised by LLM-Driven Refiner} 
    \Comment{Step 4: Adjust the edge direction if required.}
        \State $G_i \gets z' \cup G_i$
    \EndIf
    \If{$z$ is removed by LLM-Driven Refiner} 
    \Comment{Step 5: Remove the edge if deemed invalid.}
        \State $G_i \gets G_i - z'$
    \EndIf
    \If{a new edge $z''$ is added by LLM-Driven Refiner} 
    \Comment{Step 6: Add a new edge identified by the LLM.}
        \State $G_i \gets z'' \cup G_i$
    \EndIf
\EndFor
\State \textbf{Return} $G_i$ 
\end{algorithmic}
\end{algorithm}

\subsection{Causal Structure Learning}

The \textit{Causal Structure Learning} component serves as the foundational data-driven module of the ALCM framework, responsible for generating the initial causal graph from observational datasets. This component identifies causal relationships among variables by analyzing probabilistic dependencies and independencies in the data, leveraging well-established conventional causal discovery methods. These methods typically infer causal graphs by estimating relationships between variables (nodes) and their potential causal links (edges), using statistical tests to determine conditional independencies and distinguish direct relationships from those mediated by other variables. Additionally, orientation rules are applied to establish causal directions, guided by assumptions such as acyclicity or specific properties of the data distribution. Optimization techniques, including scoring functions-based methods (e.g., Bayesian Information Criterion (BIC) or likelihood-based scores), might be used to refine graph structures and select the most plausible causal model. To enhance robustness, the causal structure learning component integrates complementary methods to account for diverse data characteristics, such as linearity, non-linearity, or Gaussianity. The output of this component is an initial causal graph encapsulating key variables (nodes) and their inferred causal relationships (edges). The causal structure learning component directly influences the accuracy and reliability of both the final causal graph and future causal inferences drawn from the data.

For the implementation, we utilize three established causal discovery algorithms—Peter-Clark (PC) \cite{spirtes2000causation}, Linear Non-Gaussian Acyclic Model (LiNGAM) \cite{shimizu2006linear}, and Non-combinatorial Optimization via Trace Exponential and Augmented lagRangian for Structure learning (NOTEARS) \cite{zheng2018dags}—each chosen for their distinct strengths and complementary characteristics.

The PC algorithm leverages conditional independence (CI) tests to construct a graph's skeleton and identify v-structures, offering a robust framework for causal inference when the underlying relationships can be uncovered through probabilistic dependencies. Its ability to handle discrete and continuous variables effectively makes it a reliable choice for datasets where independence testing plays a central role. LiNGAM, on the other hand, excels at uncovering linear causal relationships in datasets with non-Gaussian distributions. By utilizing Independent Component Analysis (ICA) for causal ordering, LiNGAM demonstrates superior performance in disentangling complex linear interactions. Its focus on exploiting the statistical properties of non-Gaussianity ensures that causal directions are accurately inferred, even in the presence of latent confounders. Complementing these approaches, NOTEARS offers a novel optimization-based framework that reformulates the combinatorial problem of DAG discovery into a continuous optimization task. By incorporating an acyclicity constraint into its objective function, NOTEARS efficiently learns causal structures while maintaining scalability to larger datasets. Its gradient-based methodology makes it particularly adept at handling high-dimensional data with intricate causal dependencies.

Building on the unique strengths of these three algorithms, we propose a hybrid method that combines PC, LiNGAM, and NOTEARS within a unified framework. This hybrid approach employs a majority-weighted voting mechanism to leverage the individual advantages of each algorithm dynamically. The weighting is determined based on their relative performance on specific datasets, ensuring that the final causal graph benefits from their collective expertise. This integration enhances the robustness and reliability of the causal discovery process, allowing the hybrid method to adapt to diverse data characteristics.

The causal structure learning component synthesizes an initial causal graph by combining the outputs of these algorithms, encapsulating the potential causal linkages identified from the dataset. This graph, which represents the key variables (nodes) and their inferred causal relationships (edges), is subsequently passed to the \textit{Causal Wrapper} component for further contextualization and refinement, enabling downstream tasks to operate on a well-defined and accurate causal structure.

\subsection{Causal Wrapper}

The \textit{Causal Wrapper} component 
serves as a critical intermediary or bridge between the causal structure learning and LLM-driven refiner components. This component encapsulates and translates the raw, initial causal graph into a series of contextual, causal-aware prompts (i.e., causal prompts). These prompts are fed to the LLM-driven refiner. The primary aim of these causal prompts is to act as guides for the LLM-driven refiner, aiding it in comprehending the initial causal graph. 
Furthermore, these causal prompts direct the LLM-driven refiner to identify and integrate the relevant and updated causal background knowledge to make the solution more suited to the specific causal discovery problem at hand. Given these reasons, this prompting strategy ensures that the final causal graph
is not only precise, but also robust and reflective of the underlying causal mechanisms within the
dataset.

Equation \ref{eq:causal_prompt} shows our
causal-aware prompting strategy by infusing the context of problem and metadata information into
the prompts. This prompting strategy was inspired by an effort by Kim et al. \cite{kim2024health}. They demonstrated that contextual information is important in boosting the overall performance of LLMs' responses. 

{\small
\begin{equation}
\label{eq:causal_prompt}
Causal_{\text{prompt}} = \text{Instruction} + \text{Causal Context} + \text{Metadata} + \text{Question} + \text{Output format}
\end{equation}
}

This enhancement is accomplished by incorporating explicit elements into the prompt, with each edge being transformed into a causal prompt structured as follows:

\noindent
\textit{Instructions}: This section clarifies the role of LLMs, their objectives, and the expected behavior.
\textit{Causal Context}: It includes details about the selected causal discovery algorithm, such as its name and output.
\textit{Metadata}: This section outlines the dataset domain or variable names along with their descriptions.
\textit{Question}: It specifies the precise query, for example, whether A causes B.
\textit{Output format}: This delineates the desired format for the output.

\begin{figure}[h]
    \centering
    \includegraphics[width=1.0\textwidth]{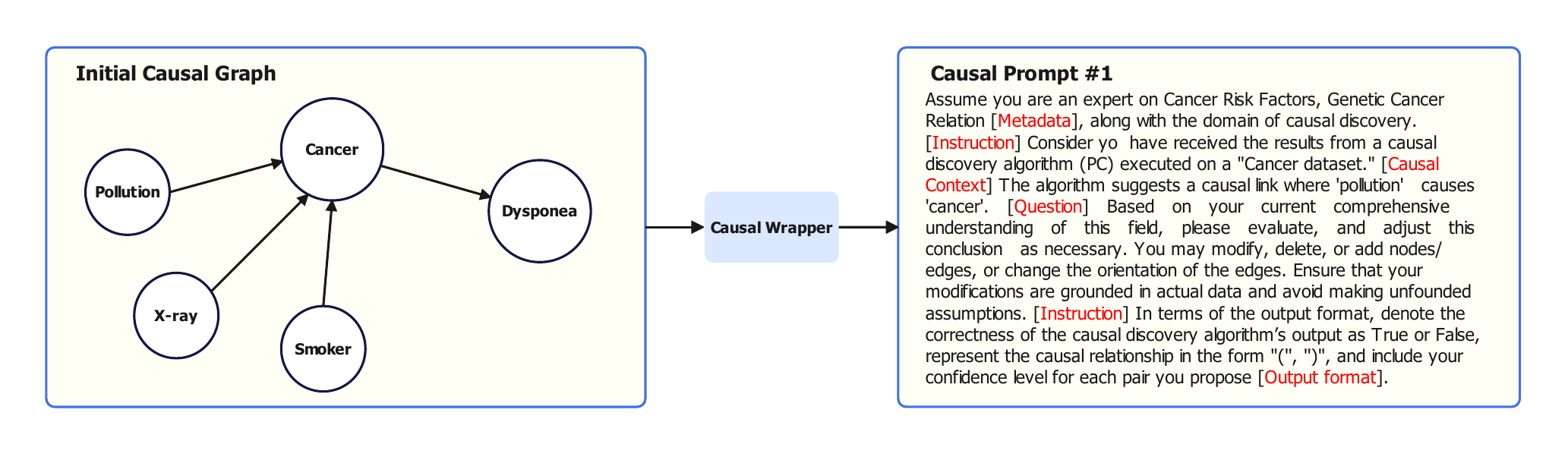}
    \caption{Causal Prompt Demonstration}
    \label{fig:Causal Prompt}
\end{figure}

Figure \ref{fig:Causal Prompt} illustrates an example of the causal wrapper's functionality. The causal structure learning component generates the initial causal graph by applying conventional causal discovery algorithms, such as Peter-Clark (PC), LiNGAM, or NOTEARS. These algorithms analyze the input observational dataset to uncover key variables (nodes) and their probabilistic dependencies, forming the skeleton of the initial graph. The nodes in this graph represent significant variables derived from the dataset, while the directed edges illustrate potential causal relationships. This foundational graph is then passed to the causal wrapper and subsequently refined by the LLM-driven refiner. Notably, the output can incorporate supplementary reasoning and confidence metrics to enhance interpretability. For instance, a simple instruction can prompt the LLM-driven refiner to engage in step-by-step reasoning using a Chain-of-Thought (CoT) approach \cite{wei2022chain}. Additionally, it can request the LLM to quantify its confidence level or provide a likelihood estimate for the generated outputs, using log-likelihood values or confidence percentages.

These prompts are critical for ensuring that the LLM comprehends the initial graph and integrates relevant causal knowledge effectively. For example, they direct the LLM to identify hidden causal relationships or validate existing edges by leveraging its contextual reasoning capabilities. This interaction facilitates the enhancement of the initial graph into a more accurate and robust representation of the underlying causal mechanisms. Once these causal prompts are generated, they are dispatched to the LLM-driven refiner component. This method ensures that the ALCM framework optimally utilizes LLMs for uncovering, refining, and validating causal relationships, thereby advancing the field of causal discovery with a high level of accuracy.

\subsection{LLM-driven Refiner}

The \textit{LLM-driven Refiner} leverages advanced language models in the refinement and evaluation of causal graphs. This component receives a series of intricately designed, contextual causal prompts from the causal wrapper component, which serve as a nuanced guide for its operations. 

The LLM-driven Refiner evaluates each edge and node in the graph by applying advanced reasoning capabilities of LLMs (e.g., GPT-4). The process involves: \begin{enumerate} \item Assessing the validity of existing causal relationships using contextual knowledge. \item Detecting and integrating hidden causal relationships by reasoning over unobserved variables. \item Reorienting or removing edges that do not align with domain knowledge or probabilistic dependencies. \item Assigning confidence scores or likelihood estimates to refined relationships, ensuring interpretability and reliability. \end{enumerate}


The LLM-driven Refiner verifies hidden causal relationships by leveraging advanced capabilities of LLMs to assess, validate, and refine the initial causal graph. This process begins with the causal prompts generated by the Causal Wrapper, which provide the LLM with explicit instructions, contextual metadata, and domain-specific information about the causal relationships. The LLM evaluates each edge in the graph based on its internal knowledge base and the provided context, determining whether the relationship is valid, needs reorientation, or should be removed. Furthermore, the LLM identifies potential hidden relationships by reasoning over unobserved variables and interactions that conventional algorithms may overlook. To ensure accuracy, the refined relationships are accompanied by confidence scores or likelihood estimates, enabling a structured and interpretable refinement process. Finally, these refined graphs are validated through cross-referencing with up-to-date domain knowledge, ensuring that the final output is both accurate and comprehensive.

The significance of the LLM-driven Refiner lies in its capacity to address and alleviate inherent limitations present in both the causal discovery algorithms and the datasets themselves. This component plays a pivotal role in uncovering and assimilating previously overlooked or concealed causal information, thereby elevating the accuracy and comprehensiveness of the causal graph. The identification and integration of hidden causal relationships into the graph are essential, as they can reveal causal connections or nodes that traditional causal discovery methods might miss or that dataset constraints could obscure.
Upon completion of the refinement process, the results are saved, and various post-processing techniques are applied to generate the final graph. These techniques involve leveraging natural language processing (NLP) to parse and extract causal relationships from textual responses provided by LLMs. Subsequently, these extracted relationships undergo validation and structuring to form a coherent causal graph.

\subsection{Interactions Between Components}

The \textit{Causal Structure Learning} component generates the initial graph, providing a data-driven foundation.
The \textit{Causal Wrapper} transforms this graph into contextualized prompts, enabling the LLM to reason about relationships in a guided and structured manner.
The \textit{LLM-driven Refiner} refines and validates the graph, identifying hidden relationships and ensuring alignment with external knowledge.
This iterative feedback loop ensures that the final causal graph is both accurate and interpretable, addressing limitations in traditional causal discovery methods and leveraging the strengths of LLMs.

\section{Implementation}

We elucidate the technical underpinnings and strategic choices behind the deployment of the ALCM framework. We provide two demonstrations of implementation of our framework to show that our framework can enhance the accuracy and generalizability.

\subsection{Implementation 1 (ALCM-PC)}

\begin{figure}[H]
    \centering
    \includegraphics[width=0.75\textwidth]{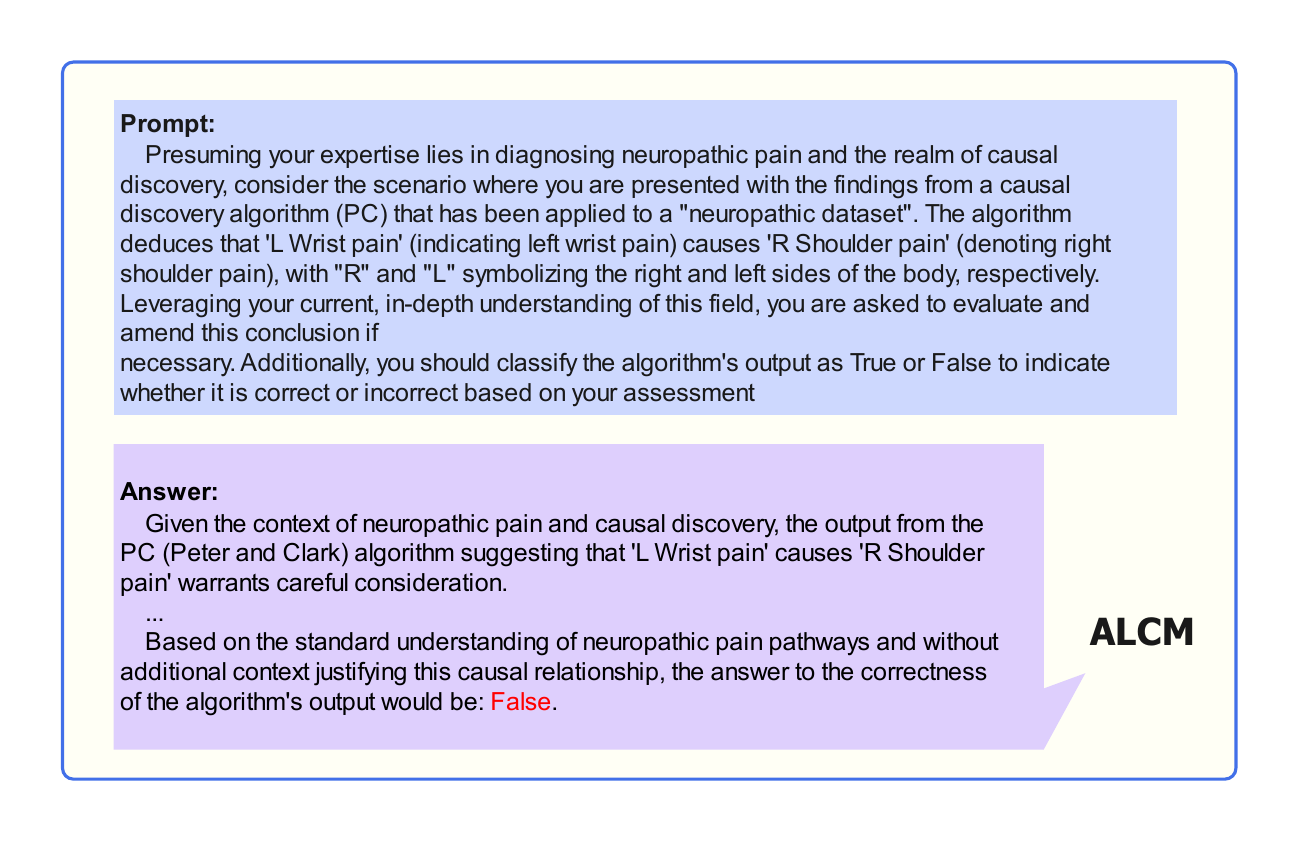}
    \caption{Prompt Template}
    \label{fig:Prompt Template}
\end{figure}

For the first implementation, we select the PC causal discovery algorithm for its robustness in handling large datasets and its efficiency in inferring causal structures through conditional independence (CI) tests. The algorithm constructs an undirected graph and iteratively removes edges by testing CI between variable pairs, conditioned on subsets of other variables, a process known as skeleton discovery. It then orients edges by detecting v-structures (triplets of nodes with specific dependency patterns) and ensuring acyclicity, resulting in a DAG that represents the causal relationships. The PC algorithm’s ability to prune unnecessary connections makes it particularly effective for high-dimensional datasets, balancing computational efficiency with accuracy. For the causal wrapper component, we utilize causal prompt. We illustrate one example of our prompt in Figure \ref{fig:Prompt Template}. For LLM-driven refiner, we exploit OpenAI GPT-4 \cite{openai2023gpt, openaiOpenAI} in our pipeline.

\subsection{Implementation 2 (ALCM-Hybrid)}

For the second implementation, we present a hybrid causal discovery approach that integrates the strengths of three leading methods: Peter-Clark (PC) \cite{spirtes2000causation}, LiNGAM \cite{shimizu2006linear}, and NOTEARS \cite{zheng2018dags}. These methods address different aspects of causal structure learning, and their combination provides a robust and accurate framework for causal discovery.

The PC method employs conditional independence (CI) tests to iteratively construct a causal graph by building its skeleton and identifying v-structures. This method is particularly effective for datasets with mixed discrete and continuous variables and excels in capturing probabilistic dependencies. Its iterative and constraint-based nature ensures computational efficiency, even in high-dimensional settings. In contrast, LiNGAM is specifically designed to uncover linear causal relationships in datasets with non-Gaussian distributions. By leveraging Independent Component Analysis (ICA), LiNGAM accurately identifies causal ordering and orients edges with high precision, even in the presence of latent confounders and linear dependencies. NOTEARS complements these approaches by reformulating causal discovery into a continuous optimization problem. By incorporating a differentiable acyclicity constraint, NOTEARS transforms the combinatorial problem of DAG discovery into a solvable optimization task, making it highly effective for datasets with intricate causal dependencies and scalable to high-dimensional data.

To leverage the unique strengths of these methods, we propose a hybrid approach that combines their outputs using dynamically assigned weights. These weights are determined based on a \textit{composite score} for each method, which captures its performance on a given dataset. The composite score is defined as the difference between the Accuracy and NHD, balancing edge-specific performance and structural alignment with the ground truth. Formally, the composite score for a method is given by:

\begin{equation}
\text{Composite}_{\text{method}} = \text{Accuracy}_{\text{method}} - \text{NHD}_{\text{method}}
\label{eq:composite_score}
\end{equation}

This score accounts for both the overall correctness of edge identification (via Accuracy) and the structural similarity of the causal graph (via NHD), ensuring that methods achieving both accurate and well-aligned graphs are given higher importance. The weights are derived by normalizing the composite scores across all methods:

\begin{equation}
W_{\text{method}} = \frac{\text{Composite}_{\text{method}}}{\sum_{\text{all methods}} \text{Composite}_{\text{method}}}
\label{eq:weights}
\end{equation}
where \( W_{\text{method}} \) represents the weight assigned to a method, ensuring that the sum of all weights equals one.

To further enhance the adaptability of the hybrid approach, we introduce a neural network-based architecture to dynamically learn these weights based on both method performance metrics and dataset-specific features. The neural network is designed to take as input the composite scores of the methods, along with features such as graph density, node degree distribution, and sparsity. Graph density quantifies how connected the graph is and is defined as the ratio of the number of edges to the maximum possible edges. Node degree distribution describes the variability in the number of connections per node, while sparsity measures the proportion of missing edges compared to a fully connected graph.

The architecture of the neural network consists of three layers:
1. An input layer with nine features, including the composite scores of the methods
(\( \text{Composite}_{\text{PC}}, \text{Composite}_{\text{LiNGAM}}, \text{Composite}_{\text{NOTEARS}} \)) and six dataset-specific features such as graph density, average node degree, and sparsity.
2. Two hidden layers with 64 and 32 neurons, respectively, each using the Rectified Linear Unit (ReLU) activation function to capture non-linear relationships among the features.
3. An output layer with three neurons (one for each method), using a softmax activation to produce normalized weights for the methods. Formally, the neural network outputs the weights as follows:

\begin{equation}
\mathbf{W} = \text{Softmax}(\mathbf{H}_2 \cdot \mathbf{W}_o + \mathbf{b}_o)
\label{eq:nn_weights}
\end{equation}
where \( \mathbf{H}_2 \) represents the outputs from the second hidden layer, \( \mathbf{W}_o \) and \( \mathbf{b}_o \) are the weights and biases of the output layer, and \( \text{Softmax} \) ensures the weights sum to one.

The neural network is trained on a dataset comprising simulated graphs with varying densities, node degrees, and sparsity levels. For each graph, the outputs of PC, LiNGAM, and NOTEARS are evaluated using Accuracy and NHD, and the composite scores are computed. The ground-truth weights for training are derived by normalizing these composite scores as described in Equation~\eqref{eq:weights}. The training objective minimizes the mean squared error (MSE) between the predicted weights and the ground-truth weights. Using the dynamically learned weights, the hybrid approach synthesizes a causal graph by aggregating the outputs of PC, LiNGAM, and NOTEARS. For each edge \( e \), the final score is computed as:

\begin{equation}
\text{Score}_{e} = \sum_{\text{method}} W_{\text{method}} \cdot \mathbb{I}_{\text{method}(e)}
\label{eq:edge_score}
\end{equation}
Here, \( \mathbb{I}_{\text{method}(e)} \) is an indicator function that equals 1 if the edge \( e \) is identified by the method and 0 otherwise. Edges with scores exceeding a predefined threshold are retained in the hybrid causal graph. For edges uniquely identified by only one method, one LLM is employed as a decisive layer. The LLM evaluates these edges based on contextual knowledge and causal reasoning to ensure that only plausible causal links are included. The validated edges are then added to the hybrid graph, enhancing its comprehensiveness and accuracy.

The resulting hybrid causal graph is passed to the \textit{Causal Wrapper} component, where it is further contextualized and refined using domain-specific templates and LLM-driven reasoning. This ensures that the final causal graph is robust, accurate, and adaptable to diverse data characteristics. By combining the strengths of traditional causal discovery methods with the adaptability of neural networks and the reasoning capabilities of LLMs, the hybrid approach achieves superior performance in causal discovery, setting a new benchmark for accuracy and robustness.

\section{Experiments}

In this section, we first present benchmark datasets used in our expermients. Next, we outline the evaluation metrics selected to measure the framework's performance in terms of accuracy, robustness, and reliability. Finally, we summarize the experimental results, demonstrating the effectiveness of the ALCM framework in generating and refining causal graphs, and its ability to reveal latent causal relationships, showcasing its advancement over existing methods.

\subsection{Benchmark Datasets}

 We utilize six benchmark datasets and their ground truth causal graphs from the BN repository: Asia, Cancer, Child, Insurance, Sachs, Sangiovese \cite{bnlearnBnlearnBayesian, long2023causal}, and also the well-known Neuropathetic dataset \cite{tu2019neuropathic} to evaluate the efficacy of the ALCM framework. These datasets are chosen for their diverse origins and complexities, covering a range of scenarios from medical studies to insurance modeling and genetic pathways. The importance of utilizing these benchmark datasets lies in their ability to provide a standardized basis for comparison, enabling the assessment of the ALCM framework's performance across varied domains and conditions. Table \ref{dataset-summary} indicates a summary of these datasets.

 \begin{table}[h]
\centering
\caption{Summary of Datasets}
\label{dataset-summary}
\begin{tabular}{|l|l|c|c|}
\hline
\textbf{Dataset} & \textbf{Domain} & \textbf{\#Nodes} & \textbf{\#Edges} \\
\hline
\text{Asia} & \text{Social Science} & 8 & 8 \\
\text{Cancer} & \text{Medical} & 11 & 18 \\
\text{Child} & \text{Social Science} & 20 & 31 \\
\text{Insurance} & \text{Finance} & 27 & 43 \\

\text{Neuropathic} & \text{Medical} & 221 & 475 \\

\text{Sachs} & \text{Biological} & 11 & 18 \\

\text{Sangiovese} & \text{Social Science} & 36 & 47 \\
\hline
\end{tabular}
\end{table}

To ensure these datasets are compatible with the input requirements of causal discovery algorithms within the ALCM framework, we implement a series of preprocessing techniques as part of the causal structure learning component. This preprocessing involves cleaning the data, handling missing values, and normalizing data formats, among other adjustments, to tailor the datasets for optimal processing. By meticulously preparing these datasets, we facilitate their effective use as inputs for the causal discovery algorithms, ensuring that the initial causal graphs generated are as accurate and informative as possible.

\subsection{Evaluation Metrics}

We select five metrics to assess the effectiveness and precision of the ALCM framework's causal discovery capabilities. The evaluation of the predicted causal graphs against the ground truth is paramount to validate the accuracy and reliability of our methodology. To this end, we employ five key metrics: precision, recall, F1-score, accuracy, and Normalized Hamming Distance (NHD), each selected for its ability to provide a comprehensive understanding of the framework's performance from different perspectives \cite{zanga2022survey}.

\begin{itemize}
    \item \textbf{Precision:} measures the proportion of correctly identified causal relationships out of all relationships identified by the algorithm. This metric is crucial for ensuring that the causal links proposed by our framework are indeed valid, minimizing false positives.

    \item \textbf{Recall:} assesses the fraction of true causal relationships that have been correctly identified by the algorithm, highlighting the framework's ability to uncover the full extent of causal connections present within the data.

    \item \textbf{F1-score:} serves as a harmonic mean of precision and recall, offering a single metric that balances both the accuracy and completeness of the identified causal relationships. This is particularly useful for comparing the overall performance of different causal discovery approaches.

    \item \textbf{Accuracy:} evaluates the overall correctness of the causal graph, including both the presence of true causal connections and the absence of false ones. This metric provides a straightforward assessment of the model's overall predictive performance.

    \item \textbf{Normalized Hamming Distance (NHD):}  quantifies the difference between the predicted causal graph and the ground truth by measuring the proportion of mismatched edges, adjusted for the size of the graph. NHD is instrumental in assessing the structural similarity of the causal graphs, offering insights into the nuanced differences that may not be captured by other metrics. In the context of a graph with \textit{m} nodes, the NHD  between the predicted graph G \textsubscript{p} and the ground-truth graph G is determined by calculating the number of edges that exist in one graph but not the other. This count is then divided by the total number of all possible edges--this formula is defined in Equation \ref{eq:1}. In essence, the NHD provides a normalized measure of dissimilarity, offering insights into the accuracy of the predicted graph compared to the ground-truth graph, accounting for the total potential edges in the graph with \textit{m} nodes.
\end{itemize}

\begin{equation} \label{eq:1}
NHD = \sum_{i=1}^{m} \sum_{j=1}^{m} \frac{1}{m^2} \cdot 1, \quad \text{where } G_{ij} \neq G_{p_{ij}}.
\end{equation}

%

\subsection{Experimental Results}

In this section, we present the experimental results and a comprehensive analysis of the ALCM framework’s performance compared to various causal discovery methods. These evaluations were conducted using seven benchmark datasets and five evaluation metrics: precision, recall, F1-score, accuracy, and Normalized Hamming Distance (NHD). The comparison encompasses traditional causal discovery methods (PC, LiNGAM, NOTEARS), a hybrid method, and LLM-based approaches. Additionally, the ALCM-PC and ALCM-Hybrid implementations, described in Sections 4.1 and 4.2, are included to demonstrate the benefits of integrating conventional algorithms with advanced refinement mechanisms.

Table \ref{your-table-label} highlights the significant improvements achieved by the ALCM framework across all datasets. ALCM-PC and ALCM-Hybrid consistently outperform other methods in precision, recall, F1-score, and accuracy, while also achieving the lowest NHD values, which indicate a closer alignment with the ground truth causal graph. ALCM-Hybrid demonstrates the highest accuracy and F1-scores across all datasets, outperforming ALCM-PC and other methods due to its ability to integrate multiple causal discovery paradigms and incorporate LLM-driven contextual refinement. ALCM-PC, which builds upon the PC method as its backbone, also performs robustly by leveraging LLM-based refinements to improve accuracy and reduce structural mismatches in the resulting causal graphs.

The experimental results show that LLM-based approaches, while capable of identifying novel causal relationships, tend to exhibit lower precision and higher NHD values. This highlights a tendency of such approaches to overgeneralize causal relationships from input data, which may result in the identification of spurious edges. Conversely, traditional methods like PC, LiNGAM, and NOTEARS demonstrate varying levels of performance. PC is effective for simpler datasets, such as Asia, where probabilistic dependencies are more straightforward to infer. However, it struggles with more complex datasets, such as Neuropathic and Sachs, which involve intricate causal dependencies. LiNGAM, tailored for linear, non-Gaussian relationships, performs well on datasets adhering to its assumptions but exhibits higher NHD values and lower precision on datasets with more diverse causal structures. NOTEARS provides scalable and efficient causal discovery but also faces limitations in capturing complex interactions in highly nonlinear datasets.

The hybrid approach (ALCM-Hybrid) capitalizes on the unique strengths of PC, LiNGAM, and NOTEARS by combining them in a weighted majority voting framework. This method dynamically assigns weights to the contributions of each algorithm based on dataset-specific characteristics, enabling it to adapt to varying causal structures. The integration of these dynamically learned weights ensures that ALCM-Hybrid achieves robust and reliable causal graph construction, as evidenced by its superior performance in metrics such as precision (up to 0.95) and accuracy (up to 98.18\%). Furthermore, the incorporation of LLMs in the ALCM framework provides an additional layer of contextual reasoning and domain-specific validation. This is particularly beneficial for datasets with intricate causal dependencies, such as Sachs and Neuropathic, where conventional algorithms alone may fail to capture the nuanced relationships between variables. By blending algorithmic rigor with AI-driven insights, the ALCM framework establishes a new benchmark for accuracy and robustness in causal discovery.

These results indicate the transformative potential of combining traditional causal discovery algorithms with LLM-driven enhancements. The ALCM framework not only addresses the limitations of existing methodologies but also demonstrates its capacity to provide reliable, interpretable, and accurate causal graphs for diverse and complex datasets.

\begin{table}[H]
\centering
\caption{Evaluation Results for Various Causal Discovery Methods}
\label{your-table-label}
\begin{tabular}{|l|l|l|l|l|l|l|}
\hline
\textbf{Dataset} & \textbf{Method} & \multicolumn{5}{c|}{\textbf{Metrics}} \\ \hline
 & & Precision & Recall & F1-Score & Accuracy & NHD \\ \hline
\multirow{7}{*}{Asia} 
 & PC & 0.75 & 0.375 & 0.5 & 33.33 & 0.1429 \\ \cline{2-7}
 & LiNGAM & 0.1818 & 0.25 & 0.2105 & 25.00 & 0.8824 \\ \cline{2-7}
 & NOTEARS & 0.1786 & 0.625 & 0.2778 & 53.57 & 0.4643 \\ \cline{2-7}
 & Hybrid & 0.452 & 0.483 & 0.466 & 47.00 & 0.193 \\ \cline{2-7}
 & LLMs & 0.1428 & 0.2174 & 0.1742 & 16.00 & 0.75 \\ \cline{2-7}
 & ALCM-PC & 1.0 & 0.5945 & 0.746 & \textbf{87.00} & \textbf{0.0893} \\ \cline{2-7}
 & ALCM-Hybrid & 0.89 & 1.0 & 0.942 & \textbf{96.6} & \textbf{0.0179} \\ \hline

\multirow{7}{*}{Cancer} 
 & PC & 0.5 & 0.5 & 0.5 & 33.33 & 0.2 \\ \cline{2-7}
 & LiNGAM & 0.5 & 0.5 & 0.5 & 50.00 & 0.6667 \\ \cline{2-7}
 & NOTEARS & 0.2 & 0.5 & 0.2857 & 50.00 & 0.5 \\ \cline{2-7}
 & Hybrid & 0.4 & 0.46 & 0.4286 & 42.00 & 0.2111 \\ \cline{2-7}
 & LLMs & 0.158 & 0.75 & 0.261 & 21.4 & 0.85 \\ \cline{2-7}
 & ALCM-PC & 0.667 & 1.00 & 0.8 & \textbf{85.71} & \textbf{0.1} \\ \cline{2-7}
 & ALCM-Hybrid & 0.9 & 0.95 & 0.924 & \textbf{90.32} & \textbf{0.0333} \\ \hline

\multirow{7}{*}{Child} 
 & PC & 0.2 & 0.28 & 0.233 & 27.00 & 0.121 \\ \cline{2-7}
 & LiNGAM & 0.14 & 0.56 & 0.224 & 56.00 & 0.8739 \\ \cline{2-7}
 & NOTEARS & 0.0474 & 0.36 & 0.0837 & 48.16 & 0.5184 \\ \cline{2-7}
 & Hybrid & 0.3 & 0.35 & 0.3231 & 34.00 & 0.2875 \\ \cline{2-7}
 & LLMs & 0.0657 & 0.48 & 0.1156 & 29.21 & 0.8765 \\ \cline{2-7}
 & ALCM-PC & 1.0 & 0.6185 & 0.764 & \textbf{78.89} & \textbf{0.047} \\ \cline{2-7}
 & ALCM-Hybrid & 0.92 & 0.85 & 0.8839 & \textbf{98.04} & \textbf{0.016} \\ \hline
 
\multirow{7}{*}{Insurance} 
 & PC & 0.2153 & 0.2692 & 0.2393 & 13.59 & 0.864 \\ \cline{2-7}
 & LiNGAM & 0.12 & 0.3462 & 0.1782 & 34.62 & 0.9022 \\ \cline{2-7}
 & NOTEARS & 0.0843 & 0.5577 & 0.1465 & 51.85 & 0.4815 \\ \cline{2-7}
 & Hybrid & 0.25 & 0.32 & 0.28 & 30.00 & 0.315 \\ \cline{2-7}
 & LLMs & 0.069 & 0.5833 & 0.1234 & 22.9 & 0.862 \\ \cline{2-7}
 & ALCM-PC & 1.0 & 0.857 & 0.923 & \textbf{94.8} & \textbf{0.054} \\ \cline{2-7}
 & ALCM-Hybrid & 0.95 & 0.91 & 0.9294 & \textbf{95.2} & \textbf{0.045} \\ \hline
\end{tabular}
\end{table}


\begin{table}[H]
\centering
\begin{tabular}{|l|l|l|l|l|l|l|}
\hline
\textbf{Dataset} & \textbf{Method} & \multicolumn{5}{c|}{\textbf{Metrics}} \\ \hline
 & & Precision & Recall & F1-Score & Accuracy & NHD \\ \hline
\multirow{5}{*}{Neuropathic} 
 & PC & 0.45 & 0.551 & 0.4954 & 51.7 & 0.1364 \\ \cline{2-7}
 & LiNGAM & 0.299 & 0.3184 & 0.3084 & 43.9 & 0.2632 \\ \cline{2-7}
 & NOTEARS & 0.5 & 0.6 & 0.55 & 55.00 & 0.1897 \\ \cline{2-7}
 & LLMs & 0.105 & 0.2831 & 0.202 & 10.2 & 0.4537 \\ \cline{2-7}
& Hybrid & 0.4355 & 0.4703 & 0.4522 & 53.21 & 0.1695 \\ \cline{2-7}
 & ALCM-PC & 0.8846 & 0.6201 & 0.7291 & \textbf{89.26} & \textbf{0.0575} \\ \cline{2-7}
 & ALCM-Hybrid & 0.89 & 0.9773 & 0.9788 & \textbf{98.18} & \textbf{0.0122} \\ \hline
 
\multirow{7}{*}{Sachs} 
 & PC & 0.4167 & 0.5882 & 0.4878 & 80.91 & 0.209 \\ \cline{2-7}
 & LiNGAM & 0.1591 & 0.4118 & 0.2295 & 41.18 & 0.5704 \\ \cline{2-7}
 & NOTEARS & 0.0303 & 0.1176 & 0.0482 & 40.15 & 0.5985 \\ \cline{2-7}
 & Hybrid & 0.3 & 0.42 & 0.351 & 43.00 & 0.2012 \\ \cline{2-7}
 & LLMs & 0.2081 & 0.6471 & 0.3149 & 63.24 & 0.9051 \\ \cline{2-7}
 & ALCM-PC & 0.6117 & 0.7059 & 0.6554 & \textbf{87.5} & \textbf{0.1881} \\ \cline{2-7}
 & ALCM-Hybrid & 0.73 & 1.0 & 0.844 & \textbf{90.00} & \textbf{0.174} \\ \hline

\multirow{7}{*}{Sangiovese} 
 & PC & 0.4348 & 0.1818 & 0.2564 & 14.71 & 0.5761 \\ \cline{2-7}
 & LiNGAM & 0.322 & 0.3455 & 0.3333 & 34.55 & 0.348 \\ \cline{2-7}
 & NOTEARS & 0.2556 & 0.6182 & 0.3617 & 55.88 & 0.2412 \\ \cline{2-7}
 & Hybrid & 0.36 & 0.47 & 0.407 & 44.06 & 0.2 \\ \cline{2-7}
 & LLMs & 0.288 & 0.6545 & 0.4 & 25.00 & 0.5143 \\ \cline{2-7}
 & ALCM-PC & 0.6548 & 1.00 & 0.7914 & \textbf{65.48} & \textbf{0.1381} \\ \cline{2-7}
 & ALCM-Hybrid & 0.87 & 0.94 & 0.903 & \textbf{93.5} & \textbf{0.065} \\ \hline
\end{tabular}
\end{table}

The results presented in Table \ref{tab:accuracy_nhd} provide a detailed evaluation of the ALCM-Hybrid framework when integrated with five prominent LLMs, namely GPT-4, Llama3.1-8B, Llama3.1-70B, Gemma2-9B, and Ministral-7B, across seven benchmark datasets. This analysis complements the broader comparison shown in Table \ref{your-table-label}, focusing specifically on the metrics of accuracy and Normalized Hamming Distance (NHD) to evaluate predictive reliability and structural alignment of causal graphs with the ground truth. As expected, GPT-4 consistently achieves the highest accuracy and lowest NHD values across all datasets, demonstrating its ability to leverage extensive pre-trained knowledge for accurate causal inference. For instance, in the Asia dataset, GPT-4 achieves an accuracy of 96.55\% with an NHD of 0.0179, significantly outperforming smaller models.


\begin{table}[H]
\centering
\caption{Accuracy and NHD Metrics for ALCM-Hybrid Across Five LLM Models}
\label{tab:accuracy_nhd}
\scriptsize
\begin{tabular}{|c|c|c|c|}
\hline
\textbf{Dataset} & \textbf{Model} & \textbf{Accuracy (\%)} & \textbf{NHD} \\
\hline
\multirow{5}{*}{Asia} & GPT-4         & 96.6 & 0.0179 \\
                      & Llama3.1-8B  & 94.32 & 0.0210 \\
                      & Llama3.1-70B & 95.10 & 0.0192 \\
                      & Gemma2-9B    & 94.80 & 0.0205 \\
                      & Ministral-7B & 93.90 & 0.0223 \\
\hline
\multirow{5}{*}{Cancer} & GPT-4         & 90.32 & 0.0333 \\
                        & Llama3.1-8B  & 88.40 & 0.0378 \\
                        & Llama3.1-70B & 89.20 & 0.0355 \\
                        & Gemma2-9B    & 88.60 & 0.0369 \\
                        & Ministral-7B & 87.50 & 0.0385 \\
\hline
\multirow{5}{*}{Child} & GPT-4         & 98.04 & 0.016 \\
                       & Llama3.1-8B  & 96.70 & 0.0205 \\
                       & Llama3.1-70B & 97.20 & 0.0193 \\
                       & Gemma2-9B    & 96.80 & 0.0201 \\
                       & Ministral-7B & 96.00 & 0.0210 \\
\hline
\multirow{5}{*}{Insurance} & GPT-4         & 95.2 & 0.045 \\
                           & Llama3.1-8B  & 94.80 & 0.0402 \\
                           & Llama3.1-70B & 95.50 & 0.0389 \\
                           & Gemma2-9B    & 94.90 & 0.0398 \\
                           & Ministral-7B & 93.70 & 0.0415 \\
\hline
\multirow{5}{*}{Sachs} & GPT-4         & 90.00 & 0.174 \\
                       & Llama3.1-8B  & 89.20 & 0.1790 \\
                       & Llama3.1-70B & 89.70 & 0.1755 \\
                       & Gemma2-9B    & 89.30 & 0.1780 \\
                       & Ministral-7B & 88.50 & 0.1805 \\
\hline
\multirow{5}{*}{Neuropathic} & GPT-4         & 98.18 & 0.0122 \\
                             & Llama3.1-8B  & 96.90 & 0.0188 \\
                             & Llama3.1-70B & 97.40 & 0.0175 \\
                             & Gemma2-9B    & 97.00 & 0.0182 \\
                             & Ministral-7B & 96.20 & 0.0193 \\
\hline
\multirow{5}{*}{Sangiovese} & GPT-4         & 93.5 & 0.065 \\
                            & Llama3.1-8B  & 91.80 & 0.0701 \\
                            & Llama3.1-70B & 92.30 & 0.0685 \\
                            & Gemma2-9B    & 91.90 & 0.0699 \\
                            & Ministral-7B & 90.70 & 0.0715 \\
\hline
\end{tabular}
\end{table}


The performance trends reveal that as model complexity and size decrease, there is a gradual decline in accuracy and an increase in NHD. Smaller models, such as Ministral-7B, while computationally efficient, exhibit limitations in capturing intricate causal dependencies, particularly in complex datasets like Cancer and Sachs. For example, in the Cancer dataset, Ministral-7B achieves an accuracy of 87.50

Llama3.1-70B and Gemma2-9B demonstrate competitive performance, approaching GPT-4’s accuracy while maintaining slightly higher NHD values, indicating room for improvement in structural fidelity. The Insurance dataset, in particular, showcases consistent performance trends across all models, suggesting that even less complex LLMs can achieve reasonable results when the dataset complexity is moderate. Despite these observations, GPT-4’s superior performance across all datasets highlights the benefits of advanced reasoning capabilities and large-scale pre-training in enhancing causal discovery tasks.

Overall, the results show the adaptability and robustness of the ALCM-Hybrid framework when paired with varying LLMs. While smaller models like Ministral-7B offer computational efficiency, they trade off precision and structural fidelity, making them less suitable for tasks requiring high accuracy. The synergy between ALCM-Hybrid and LLMs ensures robust causal inference, even when using less powerful models. These findings provide valuable insights into selecting the appropriate LLM for specific causal discovery tasks, balancing resource constraints and performance requirements. Table 4 should be placed immediately after Section 5.3 to maintain a logical flow and provide a detailed breakdown of the experimental results.

\begin{figure}[H]
    \centering
    \begin{subfigure}[b]{0.45\textwidth}
        \centering
        \includegraphics[width=\textwidth]{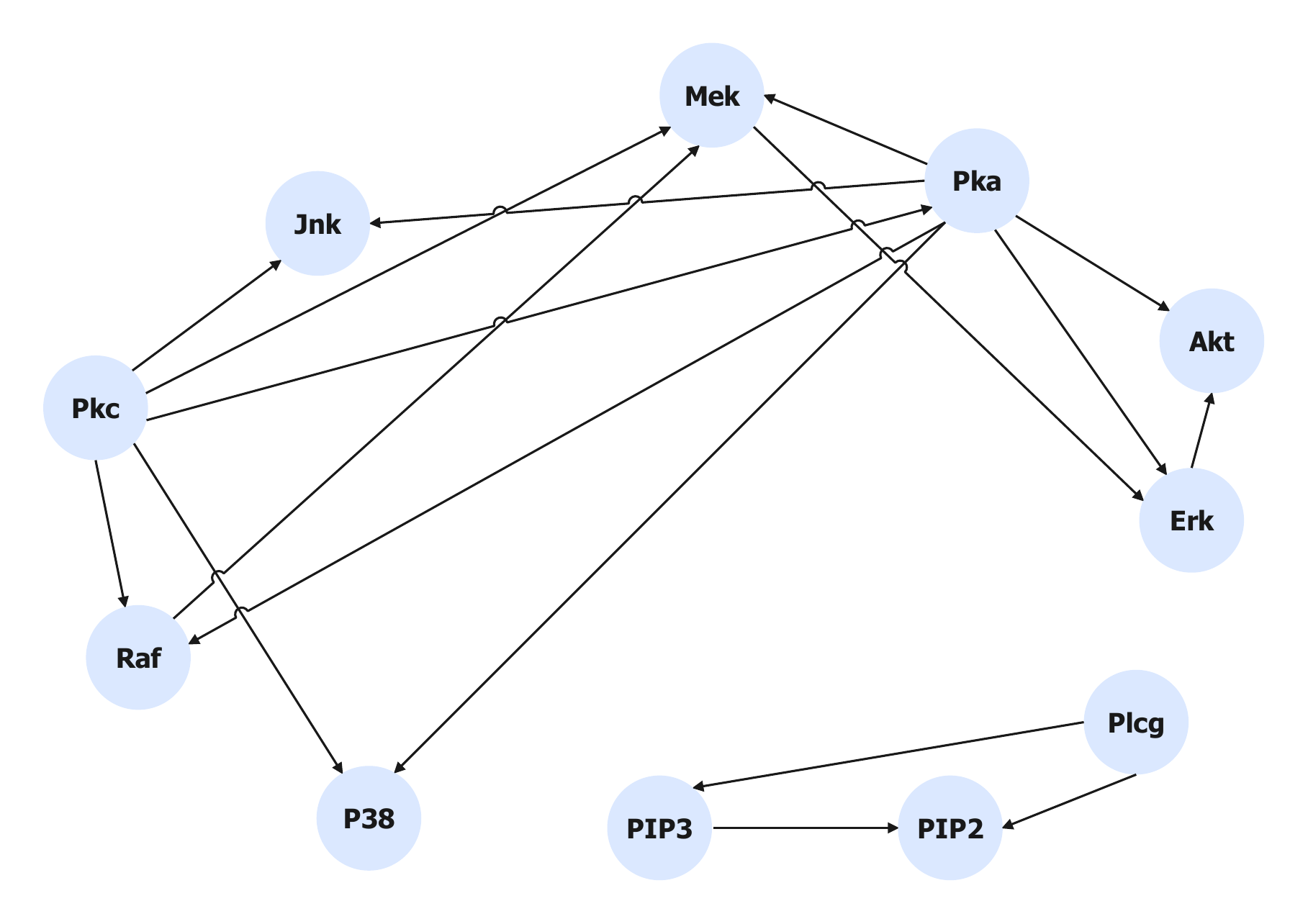}
        \caption{Causal graph for Ground Truth}
        \label{fig:gt}
    \end{subfigure}
    \hspace{10pt}
    \begin{subfigure}[b]{0.43\textwidth}
        \centering
        \includegraphics[width=\textwidth]{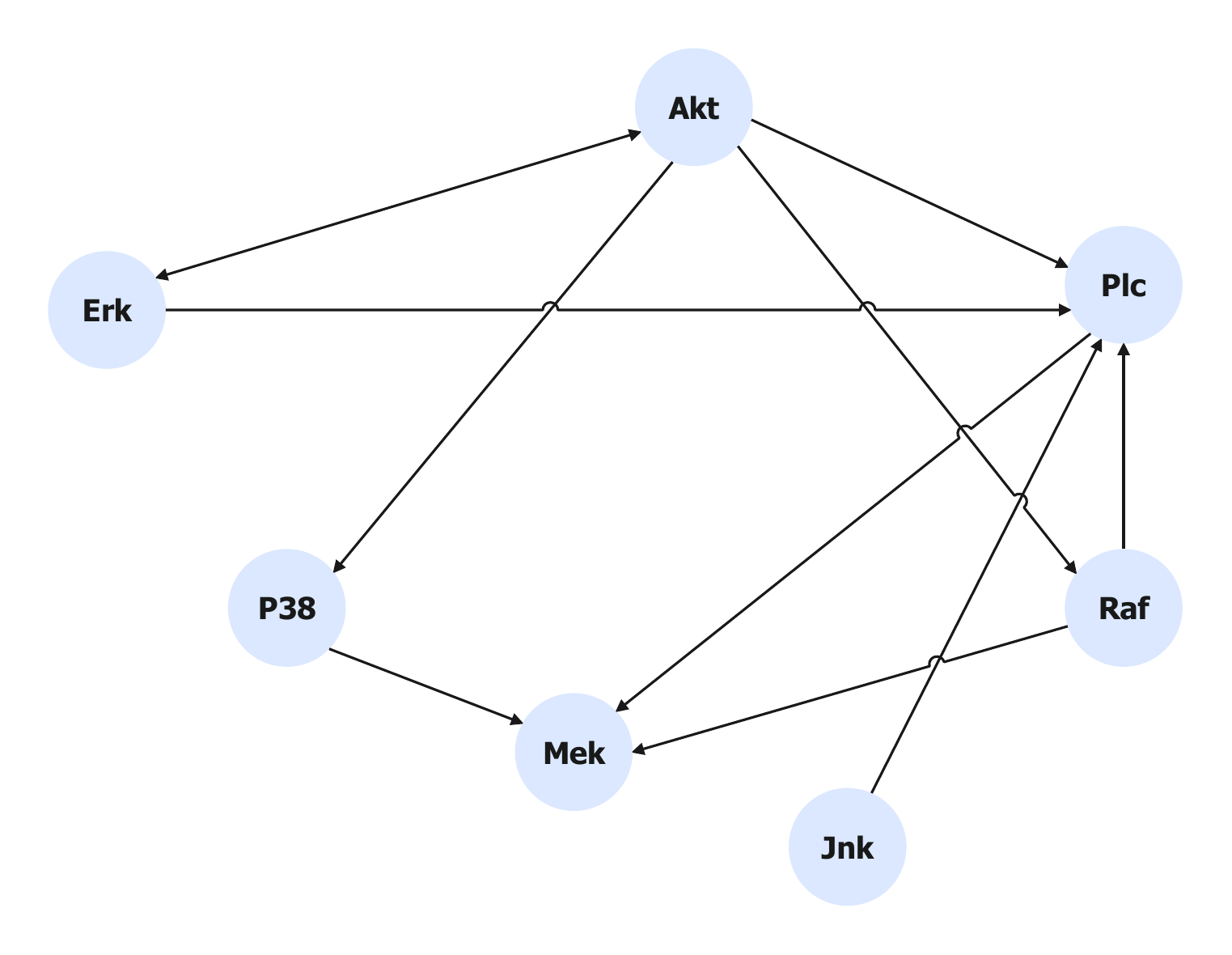}
        \caption{Causal graph for LLMs-based}
        \label{fig:llm}
    \end{subfigure}
    \hfill 
    \begin{subfigure}[b]{0.53\textwidth}
        \centering
        \includegraphics[width=\textwidth]{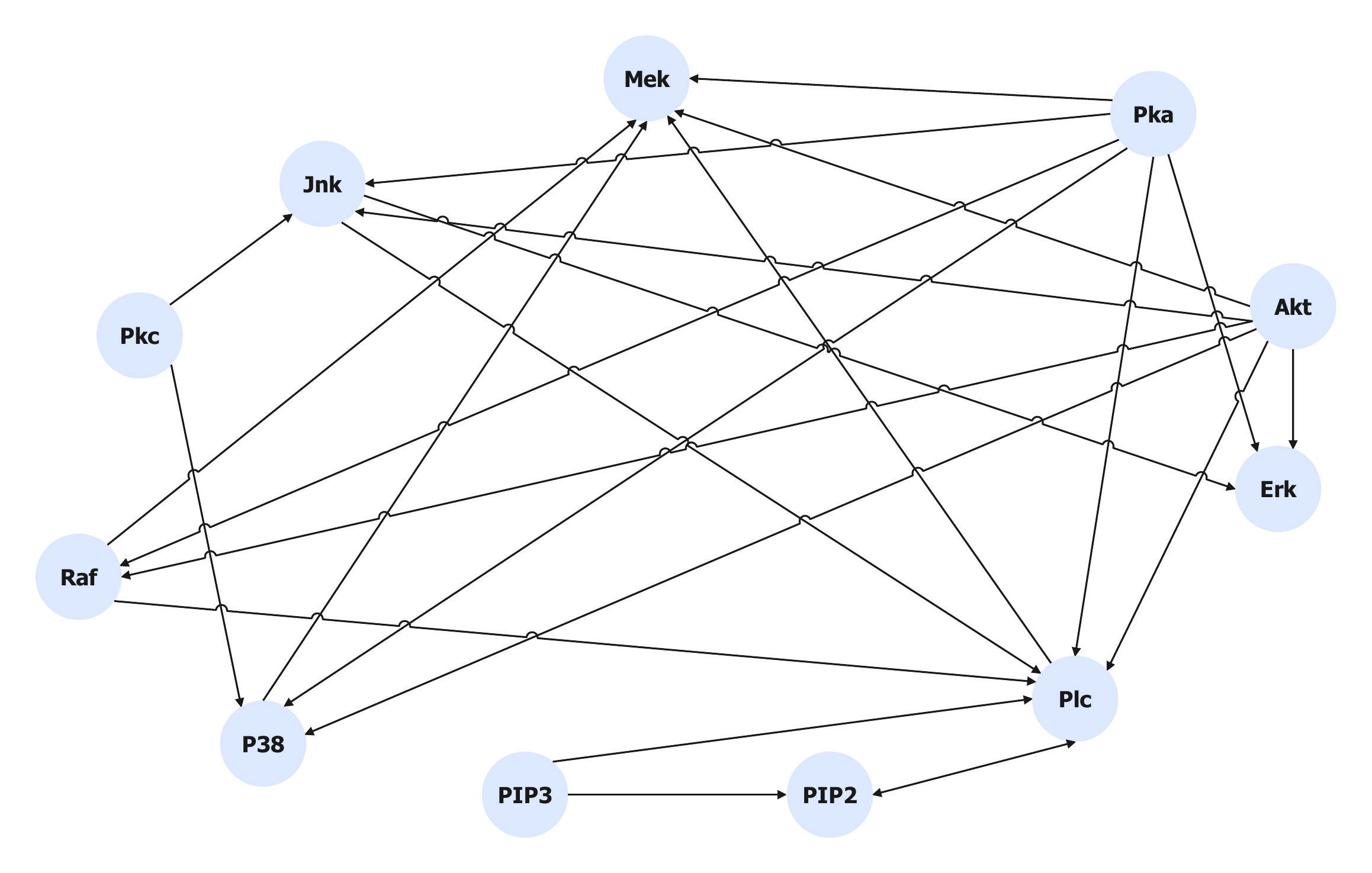}
        \caption{Causal graph for PC}
        \label{fig:pc}
    \end{subfigure}
    
    \vspace{10pt}
    
    \begin{subfigure}[b]{0.47\textwidth}
        \centering
        \includegraphics[width=\textwidth]{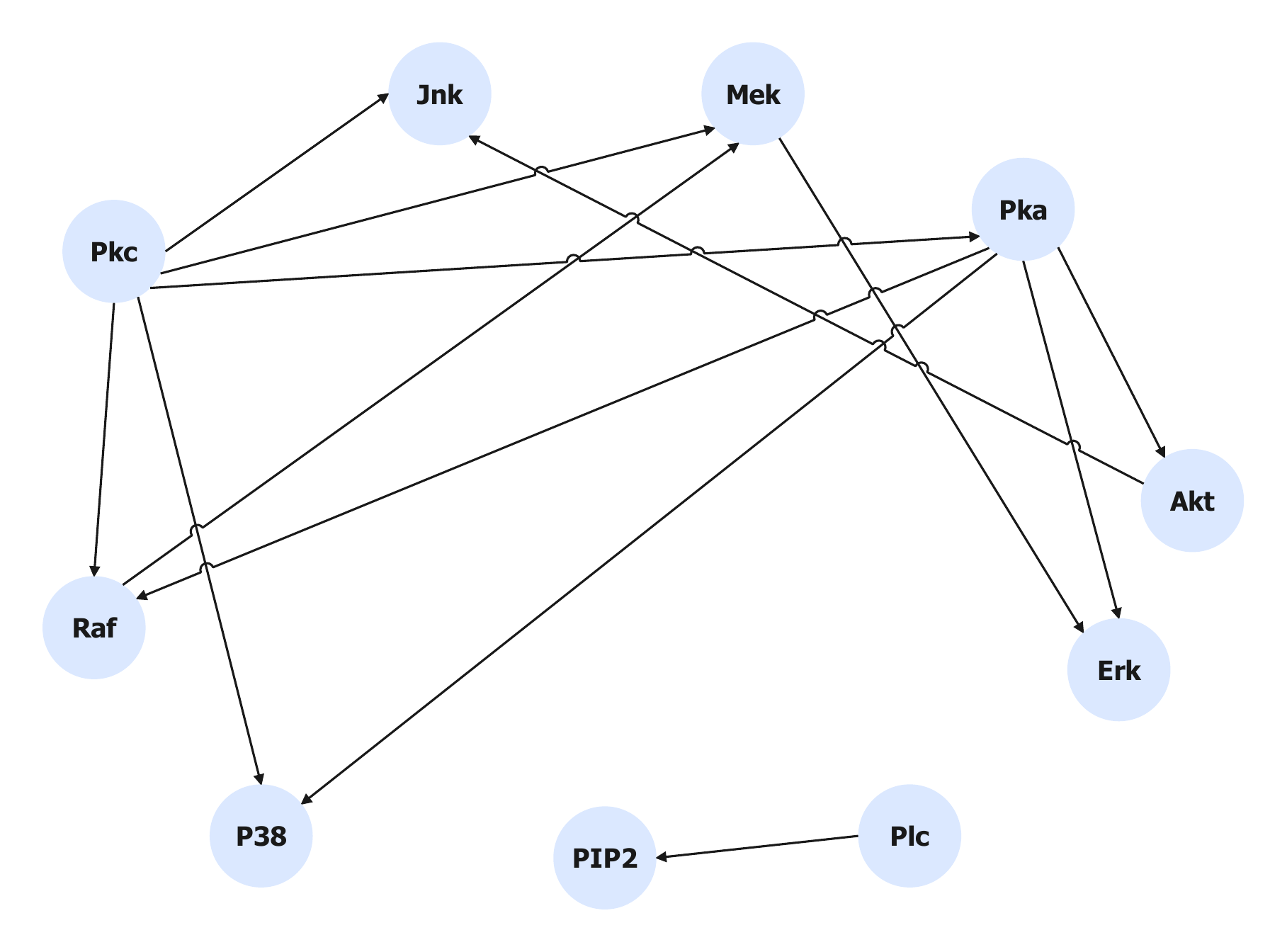}
        \caption{Causal graph for ALCM}
        \label{fig:alcm}
    \end{subfigure}
    \hspace{10pt}
    \begin{subfigure}[b]{0.47\textwidth}
        \centering
        \includegraphics[width=\textwidth]{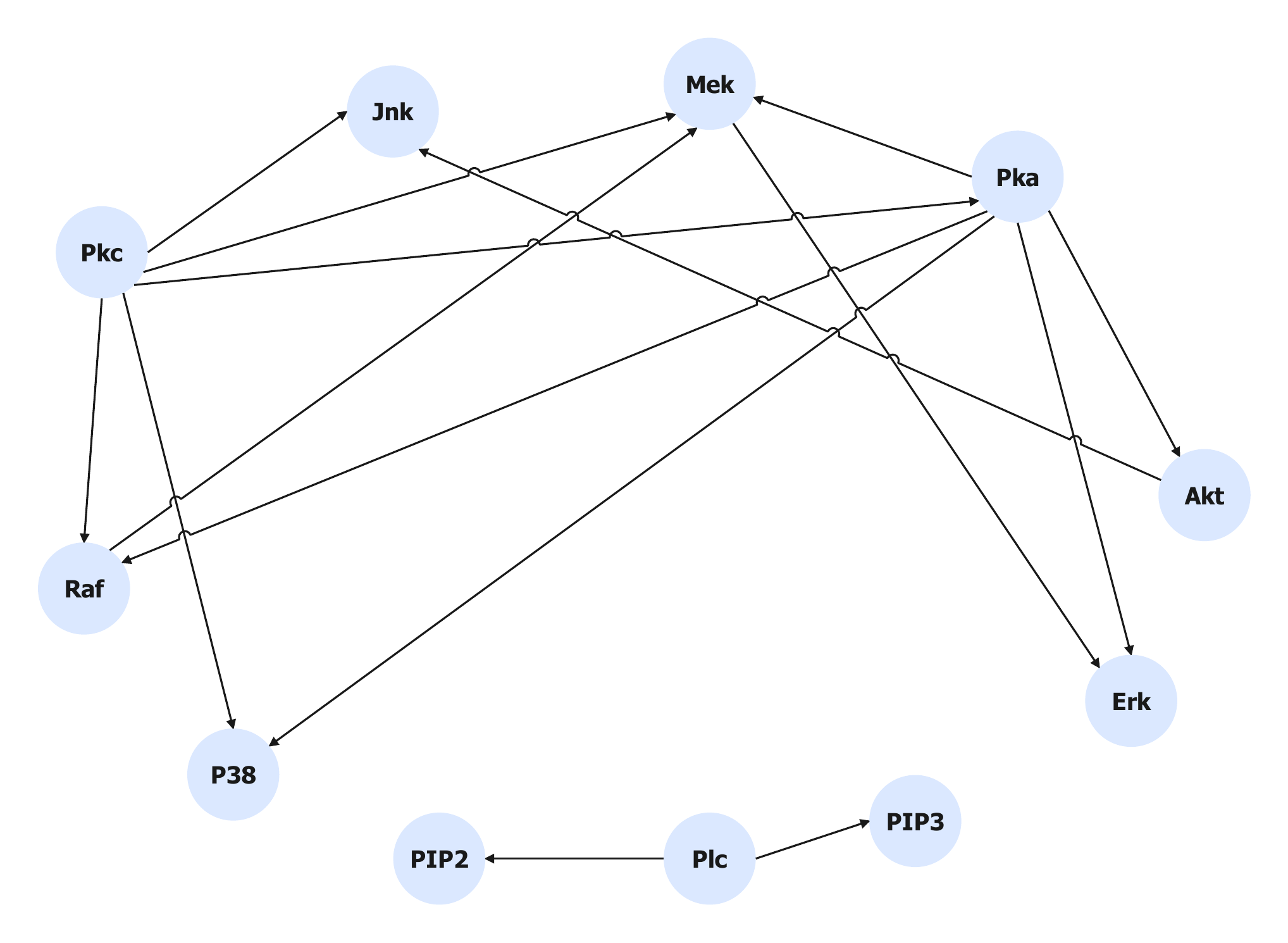}
        \caption{Causal graph for ALCM-Hybrid}
        \label{fig:alcm-hybrid}
    \end{subfigure}
    \caption{Causal graphs Demonstrations}
    \label{fig:demonstrations}
\end{figure}

We depict the causal graphs obtained by a couple of causal discovery methods on Sachs dataset in Figure \ref{fig:demonstrations}. The Sachs dataset \cite{bnlearnBnlearnBayesian} includes data on 11 phosphorylated proteins and phospholipids from human immune cells, providing a basis for analyzing protein signaling pathways and constructing causal networks. It is especially valuable for causal discovery research, with data collected from cells under different experimental conditions, making it an excellent benchmark for testing causal discovery algorithms. Graph of ground truth, LLMs-based approach, PC, ALCM, ALCM-Hybrid are shown in Figures \ref{fig:gt}, \ref{fig:llm}, \ref{fig:pc}, \ref{fig:alcm}, \ref{fig:alcm-hybrid}, respectively.

\begin{figure}[th!]
    \centering
    \includegraphics[width=0.7\textwidth]{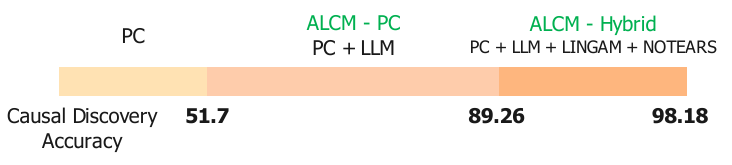}
    \caption{Additive Contribution on Causal Discovery Accuracy on Neuropathetic Pain}
    \label{fig:additive contribution}
\end{figure}

\begin{figure}[th!]
    \centering
    \includegraphics[width=0.7\textwidth]{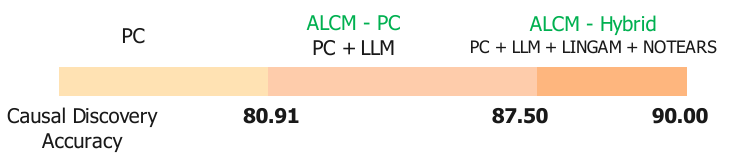}
    \caption{Additive Contribution on Causal Discovery Accuracy on Sachs}
    \label{fig:additive contribution on Sachs}
\end{figure}

The enhanced performance across all metrics for both ALCM-PC and ALCM-Hybrid variants can be directly linked to their innovative methodologies. ALCM's use of LLMs introduces a layer of causal reasoning and validation that is absent in traditional approaches, while the hybrid model further capitalizes on this by combining algorithmic precision with AI's contextual insights. This strategic amalgamation ensures that our framework is at the forefront of causal discovery, setting a new benchmark for accuracy, comprehensiveness, and applicability in the field. We also visualize the additive
contributions of each causal discovery framework in Figures \ref{fig:additive contribution} and \ref{fig:additive contribution on Sachs} on two benchmarks--neuropathetic and sachs.

\subsection{Results for Adding New Nodes or Edges}

\begin{figure}[ht]
     \centering
     \begin{subfigure}[t]{0.35\linewidth}
         \centering
         \includegraphics[width=\linewidth]{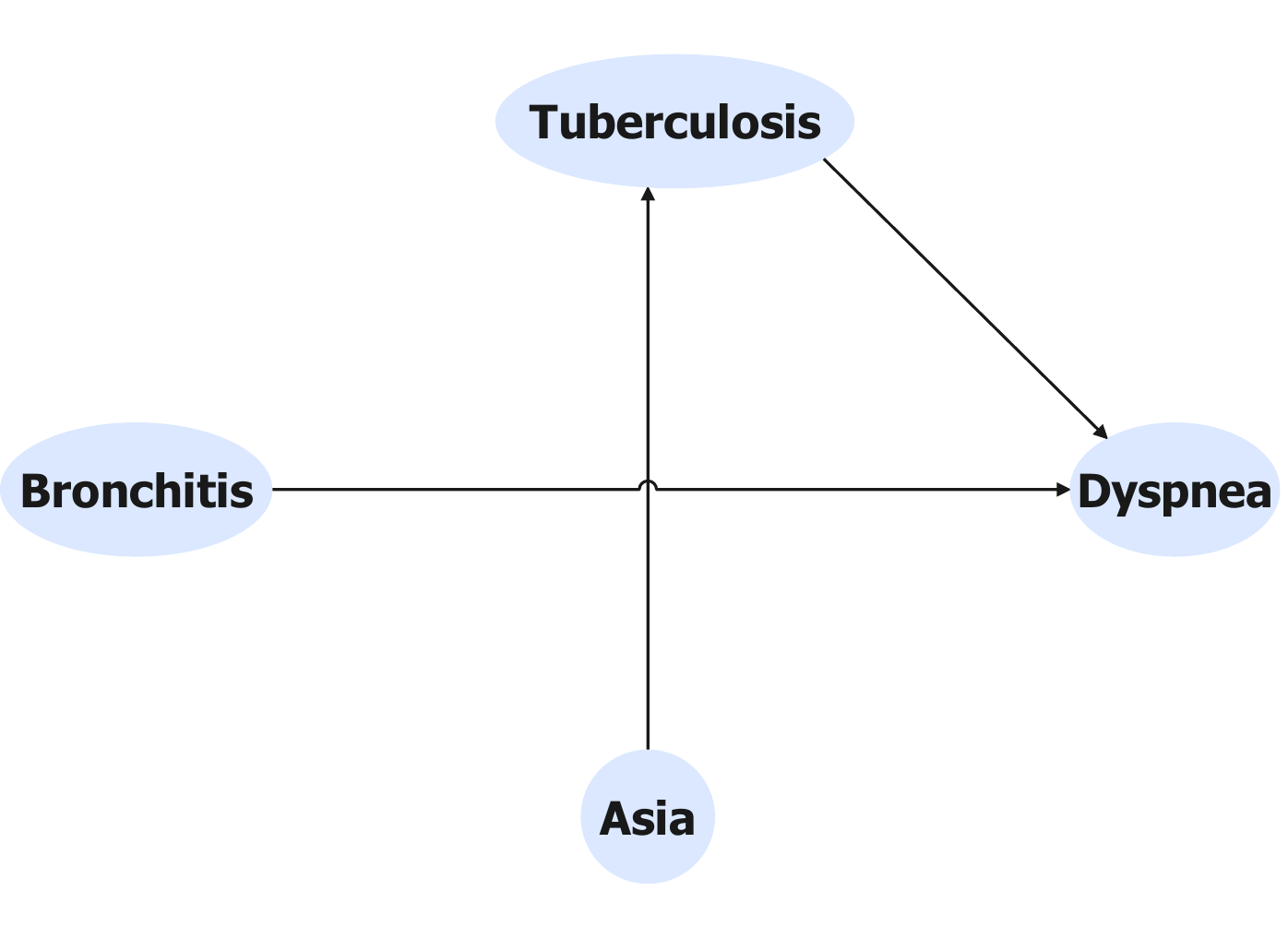}
         \caption{Causal graph for PC}
         \label{fig: Causal graph for PC}
     \end{subfigure}
     \hspace{40pt} 
     \begin{subfigure}[t]{0.45\linewidth}
         \centering
         \includegraphics[width=\linewidth]{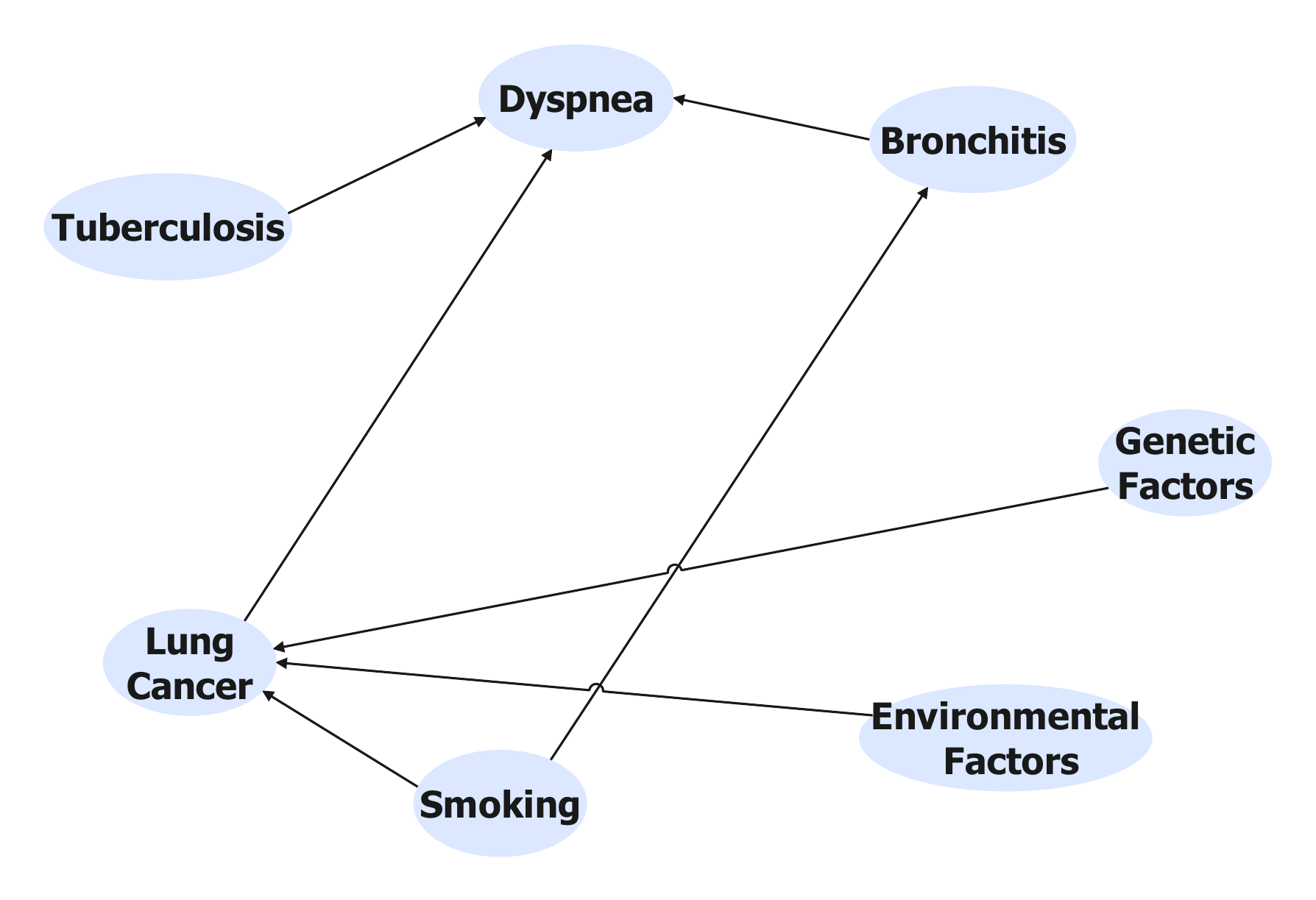}
         \caption{Causal graph for ALCM}
         \label{fig:Causal graph for ALCM}
     \end{subfigure}
     \caption{Causal graphs for demonstrating new nodes or edges}
\end{figure}

The extensive updated knowledge and expert supervision provided by LLMs significantly facilitate the identification of elusive variables (Markov blanket) and causal connections. These might remain undetected or in the dataset or overlooked by causal discovery algorithms. 
Figure\ref{fig: Causal graph for PC} and \ref{fig:Causal graph for ALCM} show these capability of unmasking these hidden aspects.  As Figure \ref{fig: Causal graph for PC} demonstrates, the causal discovery algorithm (PC) fails to detect all of the true nodes and edges, but ALCM can provide us with new nodes or edges that not present in the output set of causal discovery algorithm as illustrated in \ref{fig:Causal graph for ALCM}. We prompted LLMs to provide us the confidence level for its responses as well. The validity of ALCM answer is also confirmed by the up-to-the-date medical articles, including \cite{iarc2004tobacco}.


\begin{figure}[ht!]
  \centering
  \includegraphics[trim=0cm 18cm 0cm 0cm, clip, width=\textwidth]{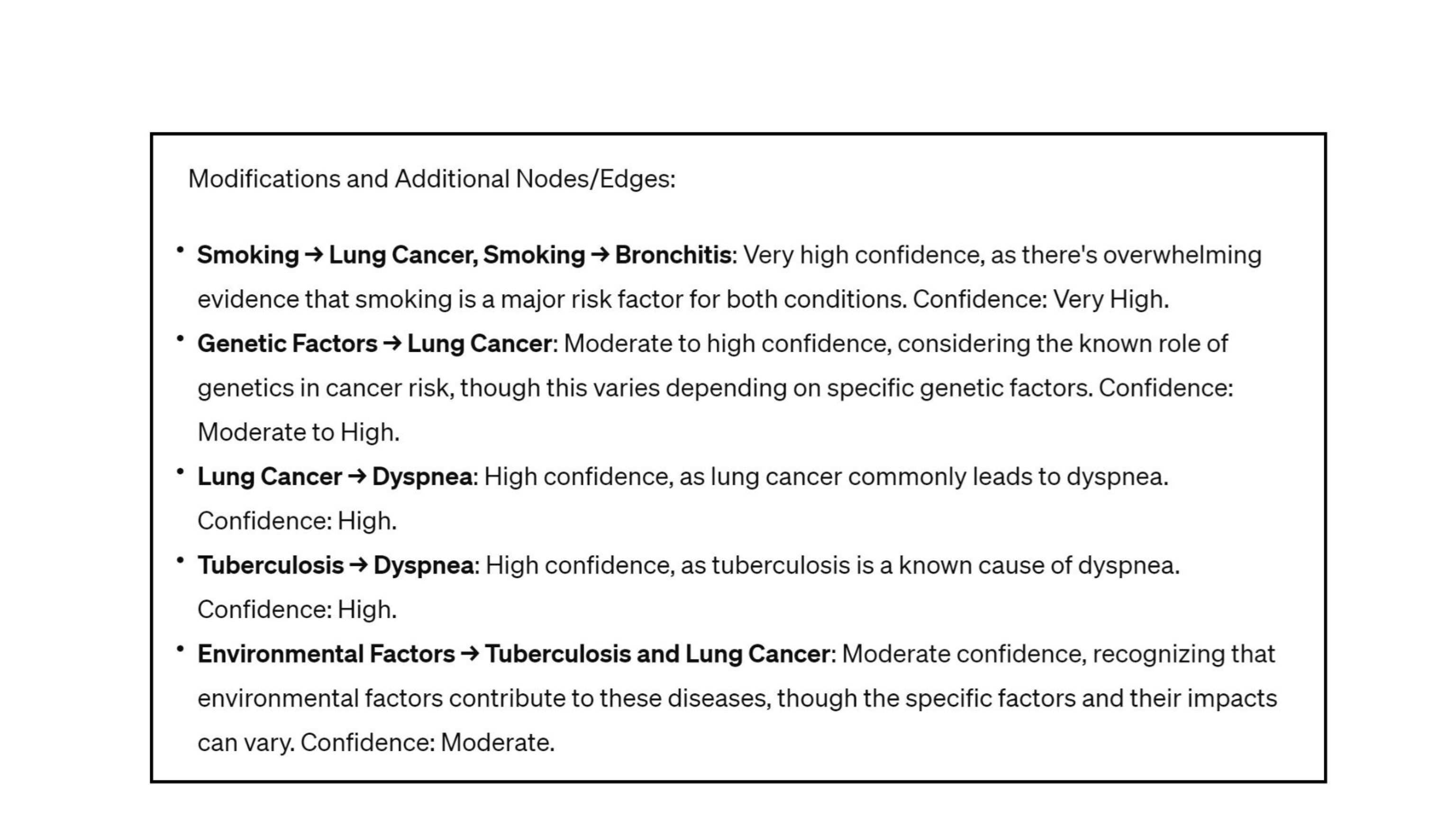}
  \captionsetup{skip=1pt} 
  \caption{Results for Uncover, Hidden, or Ignored Nodes and Edges}
  \label{fig:Uncover Hidden or Ignored Nodes and Edges}
\end{figure}

The traditional causal discovery depends on the structured dataset and their quality which are curated and annotated  by human experts. However, these dataset are neither available in a wide range of domains or can be generalize to the new tasks. Hence, we empower ALCM by viture of LLMs component with this capability to uncover hidden variables and causal connections. Figure~\ref{fig:Uncover Hidden or Ignored Nodes and Edges} indicates the ALCM capability to entangle the hidden variables and causal relations which are not present in the dataset.

Figure \ref{fig:Uncover Hidden or Ignored Nodes and Edges} illustrates the confidence levels associated with the causal relationships refined or discovered by the ALCM framework. These confidence levels are provided by the LLM-driven refiner and represent the degree of certainty for each modification or addition to the causal graph. For instance, in Figure 8, confidence scores are categorized as "Very High," "High," or "Moderate," reflecting the LLM's assessment based on domain knowledge and available data. These scores play a critical role in validating the plausibility of newly added nodes and edges, such as "Smoking → Lung Cancer" and "Environmental Factors → Tuberculosis and Lung Cancer," which traditional causal discovery algorithms might overlook. The use of confidence levels ensures that the refined causal graph not only aligns with existing domain knowledge but also achieves a higher level of interpretability and reliability.


\section{Limitations}

While the ALCM framework demonstrates significant advancements in leveraging LLMs for causal discovery, several limitations need to be addressed. First, the framework's reliance on LLMs introduces challenges related to the accuracy and reliability of the generated causal refinements, as LLMs can occasionally generate incorrect or hallucinated outputs. This limitation highlights the importance of robust validation mechanisms, such as cross-referencing with domain-specific knowledge and incorporating human oversight where necessary.

Second, the scalability of the ALCM framework to large, high-dimensional datasets remains constrained by the computational costs associated with running LLMs and the underlying causal discovery algorithms. Optimizing the framework for computational efficiency is essential for broader applicability in real-world scenarios.

Third, the framework assumes the availability of high-quality input datasets and metadata, which might not always be accessible in certain domains. In cases of sparse or biased data, the effectiveness of ALCM's causal graph generation may be compromised.

Last, the framework currently lacks a mechanism to quantify uncertainties in the generated causal graphs beyond the confidence scores provided by the LLMs. Future iterations of the framework could integrate statistical measures to provide more comprehensive uncertainty quantification.

\section{Future Work}

In the subsequent phases of our research, we aim to develop a more sophisticated causal-aware framework. This framework will leverage the power of knowledge graphs, which are instrumental in augmenting the accuracy of our models. 
Furthermore, we plan to explore the integration of our framework with Monte Carlo Tree Search (MCTS). This integration is envisioned to evolve our system into a more dynamic and adaptive problem-solving agent. 


Additionally, to advance the ALCM framework's capabilities and address the issue of LLM hallucination, we propose integrating ALCM with the Retrieval-Augmented Generation (RAG) system and openCHA \cite{abbasian2023conversational}. This integration aims to harness RAG's ability to augment LLMs' generative processes with data retrieval, ensuring that causal discovery are grounded in relevant and factual information. openCHA sophisticated dialogue capabilities will further enhance ALCM by enabling dynamic, interactive validation of causal hypotheses. 


\section{Conclusion}

This study showed the transformative potential of combining LLMs with data-driven causal discovery algorithms through the introduction of the Autonomous LLM-Augmented Causal Discovery Framework (ALCM). The ALCM emerges as a groundbreaking solution, aiming to enhance the generation of causal graphs by leveraging the sophisticated capabilities of LLMs alongside conventional causal discovery techniques. By integrating causal structure learning, a causal wrapper, and an LLM-driven causal refiner, ALCM facilitated an autonomous approach to causal discovery, significantly outperforming existing methodologies in both accuracy and interpretability. The empirical validation of ALCM not only attests to its superior efficacy over prevailing LLM methods and conventional causal reasoning mechanisms but also illuminates new pathways for leveraging LLMs in uncovering intricate causal relationships across a myriad of domains. 



\bibliographystyle{plain}
\bibliography{references}

\end{document}